\documentclass{article}

\usepackage{microtype}
\usepackage{graphicx}
\usepackage{subfigure}
\usepackage{booktabs} 
\usepackage{amsmath}
\usepackage{amssymb}
\usepackage{bm}

\usepackage{hyperref}

\usepackage[accepted]{mlsys2025}

\makeatletter
\renewcommand{\mlsys@appearing}{}
\makeatother
\hypersetup{pdfsubject={}}

\mlsystitlerunning{FastTPS: An Optimized Method for LLM Token Phase for AI accelerators}

\begin{document}

\twocolumn[
\mlsystitle{FastTPS: An Optimized Method for LLM Token Phase for AI accelerators}



\mlsyssetsymbol{equal}{*}

\begin{mlsysauthorlist}
\mlsysauthor{Wenzong Yang}{amd}
\mlsysauthor{Danyang Zhang}{amd}
\mlsysauthor{Kun Cao}{amd}
\mlsysauthor{Siddagangaiah, Tejus}{amd}
\mlsysauthor{Patwari, Rajeev}{amd}
\mlsysauthor{Zhanxing Pu}{amd}
\mlsysauthor{Siyin Kong}{amd}
\mlsysauthor{Zijiang Yang}{amd}
\mlsysauthor{Hao Zhu}{amd}
\mlsysauthor{Sharma, Varun}{amd}
\mlsysauthor{Yue Gao}{amd}
\mlsysauthor{Tianping Li}{amd}
\mlsysauthor{Fan Yang}{amd}
\mlsysauthor{Jicheng Chen}{amd}
\mlsysauthor{Yushan Chen}{amd}
\mlsysauthor{Fennian Zhao}{amd}
\mlsysauthor{Ng, Aaron}{amd}
\mlsysauthor{Delaye, Elliott}{amd}
\mlsysauthor{Sirasao, Ashish}{amd}
\mlsysauthor{Nag, Sudip}{amd}
\end{mlsysauthorlist}

\mlsysaffiliation{amd}{Advanced Micro Devices, Inc., Santa Clara, California, USA}

\mlsyscorrespondingauthor{Wenzong Yang}{booker.yang@amd.com}
\mlsyscorrespondingauthor{Danyang Zhang}{Danyang.Zhang@amd.com}

\mlsyskeywords{Machine Learning, MLSys}

\vskip 0.3in

\begin{abstract}
The popularity of large language models (LLMs) escalates an ongoing demand for effective inference. However, due to the sequential processing of tokens during the token phase in decoder-only LLMs inference, the inherent low parallelism leads to reduced throughput and suboptimal utilization of the computing units on artificial intelligence (AI) accelerators, particularly when handling long-sequence inputs that impose significant memory overhead. Recently, many reported methods have been developed as potential solutions, since they emerge with numeric deviation. This paper presents FastTPS, a high performance and low-precision loss method for accelerating the token-phase in LLM inference on general AI accelerators which includes three key components: (1) AI accelerator-enabled reloading-free KV Cache concatenation which decreases memory access overhead as well as enables full fusion of Attention, (2) high-efficiency and high-accuracy ‘RoPE’  attention based on the tiling optimized FLAT, and (3) highly-fused MLP with fine-grain pipeline scheduling. Our results confirm that FastTPS significantly alleviates memory bottlenecks in the token phase, delivering a 6× speed improvement (compared to none-fusion) on an AMD Ryzen AI 300 series NPU with BF16 precision while sustaining 93\% peak memory bandwidth utilization during Phi3-mini-4k-instruct inference.
\end{abstract}
]



\printAffiliationsAndNotice{}  

\section{INTRODUCTION}
\label{introduction}

The remarkable performance of large language models (LLMs) has been widely demonstrated across diverse applications, including content generation, machine translation, text classification, and conversational question answering~\cite{ref1,ref2,ref3,ref4,ref5,ref6}. LLMs have been extensively deployed across cloud-server clusters, edge platforms, and hybrid edge-server infrastructures~\cite{ref7,ref8,ref9,ref10,ref11,ref12,ref13,ref14}. The context length of LLMs has been continuously extended to accommodate emerging use cases such as high-resolution image understanding and code generation. This growth leads to a quadratic increase in both computational complexity and memory requirements. However, the hardware development cycle has struggled to keep pace with the rapid scaling of Transformer model sizes. Consequently, optimizing Transformer inference on AI accelerators with constrained computational and memory resources has emerged as a critical research challenge.

Transformers, especially the decoder-only architecture, form the foundational framework for the majority of contemporary LLMs~\cite{ref15}. The inference process of a decoder-only Transformer consists of two distinct phases: the prefill phase and the token phase. Among them, the token phase processes only a single token per iteration due to its autoregressive dependency. This sequential nature severely limits computational resource utilization under bandwidth constraints, regardless of whether inference is performed on single-cluster or multi-cluster GPU architectures or edge accelerators~\cite{ref16}. Furthermore, the KV (key-value) Cache is a key component of LLM inference which requires frequent load-store operations causing a serious shortage of bandwidth~\cite{ref6}. Although some quantization KV Cache methods can significantly reduce this overhead, a large amount of memory read and write still exists~\cite{ref17}.

As mentioned in~\cite{ref18}, the Transformer architecture fundamentally comprises two distinct computational components: attention and multi-layer perceptions (MLPs). Emerging token phase optimization techniques for attention mechanisms prominently feature model compression methodologies and numerical acceleration paradigms typified by FlashAttention~\cite{ref18,ref19,ref20,ref21,ref22,ref23,ref24}. Although these approaches can effectively curtail computational cost, their practical applicability is often constrained by quantization-induced accuracy losses and auxiliary processing burdens~\cite{ref20,ref21,ref25}, precluding deployment in high-precision scenarios~\cite{ref26,ref27,ref28}. Meanwhile, the MLP submodule constitutes 66.7\% of Transformer parameters~\cite{ref18}, serving as the computational cornerstone of LLM performance. Despite advances in the MLP architecture simplification through component fusion~\cite{ref29}, the systematic token phase optimization targeting MLP-specific computations remains remarkably underdeveloped in current literature.

This paper proposes a dataflow optimization strategy, termed fast tokens per second (FastTPS), specially designed for general-purpose AI accelerators to fundamentally address the challenges associated with the token phase in LLM inference. It focus on the following difficulties that need to be solved in the token phase optimization: (1) low memory utilization: the KV Cache concatenation process requires reloading of previously stored key-value pairs for each newly generated token which substantially reduces the memory efficiency of the maintained cache structure; (2) tight hardware resources: the limited parallelism of small batching in the token phase restricts the processor’s ability to fully exploit hardware computational resources under the constrained memory bandwidth conditions; (3) high accuracy requirements: on the one hand, some applications~\cite{ref26,ref27,ref28} are sensitive to inference accuracy. On the other hand, most LLM models have been integrated with quantization~\cite{ref18,ref19}, pruning~\cite{ref20,ref21} and other LLM architecture optimization methods. It makes the following key contributions:
\begin{enumerate}
    \item \textbf{Global KV Cache management (GKVC).} In this paper, we propose a global 3D static KV Cache memory management strategy for token phase which updates KV to reserved location gradually by the AI accelerator or software. Consequently, GKVC reduces the number of data transfers, alleviates bandwidth pressure, and eliminates the need for memory reorganization with ‘Concat’ operator. At the same time, it ensures the continuity of the TPS process on AI accelerator which is convenient for subsequent operations’ fusion and workflow arrangement. 
    \item \textbf{Fused logit attention tiling for TPS (TPSFLAT).} Building upon the FLAT~\cite{ref30} concept, this paper introduces TPSFLAT, an optimization rotary positional embedding (‘RoPE’) attention scheme tailored for the TPS computation process. Without compromising computational accuracy, TPSFLAT fuses a greater number of operators based on GKVC by taking into account the specific characteristics of the TPS workflow to enhance the Operational Intensity (OI) further. In addition, to address the limitations imposed by Amdahl’s Law, we design a data tiling strategy aligned with a three-level cache mechanism, aiming to improve the parallelism of computational tasks. 
    \item \textbf{Multi-op fusion method based on interlaced layout of gate-up weights (Fusion MLP). } In the MLP block, we design an interlaced layout of gate-up weights and implement a new multi-op fusion scheduling method. That provides necessary conditions for subsequent multi-op fusion and scheduling through fine-grained matrix division and rearrangement. This approach greatly reduces the I/O overhead and improves the hardware utilization.
\end{enumerate}

Based on the AMD Ryzen AI 300 series NPU AI accelerator~\cite{ref31,ref32,ref33}, we evaluate the effectiveness of the FastTPS scheme across several state-of-the-art LLMs. Experimental results demonstrate that, compared to the standard implementations of LLMs in HuggingFace~\cite{ref38}, FastTPS achieves more than 5x speedup. Top-down analysis shows that it introduces targeted optimizations for two critical components in LLMs: the attention block and the MLP block. They are accelerated up to 13.07x and 3.14x respectively. In terms of numerical precision, FastTPS achieves up to 10x higher accuracy compared to FlashAttention. This suggests it can preserve high computational precision while accelerating the LLM inference. Our research demonstrates that, although the LLM token generation is inherently constrained by sequential processing, FastTPS mitigates bandwidth underutilization and approaches or increases theoretical performance limitation through the optimized KV Cache management and workflow.

\section{BACKGROUND}

\subsection{Terminology}

\begin{figure}[htbp]
    \vspace{-0.05in}
    \centering
    \includegraphics[width=0.5\textwidth]{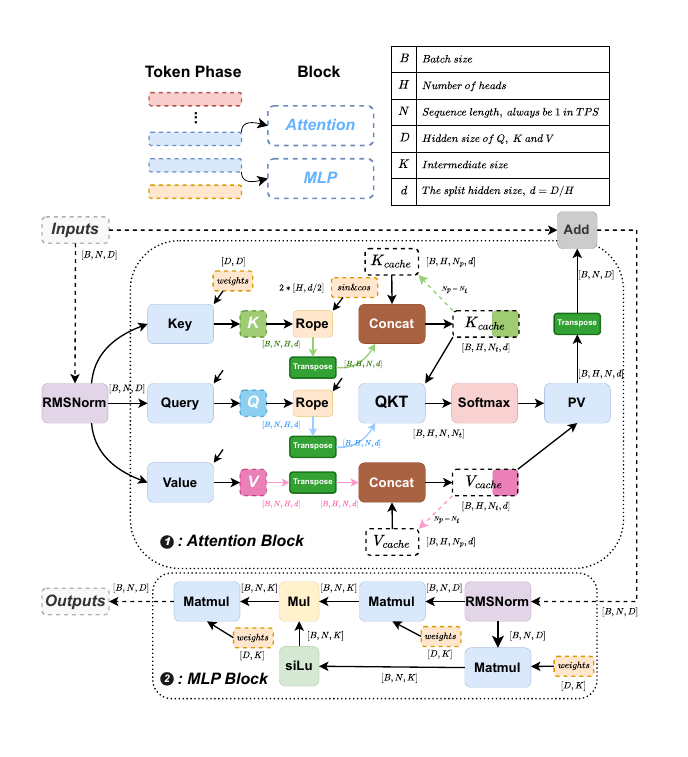}
    \vspace{-30pt}
    \caption{\small Structural overview of the standard attention and MLP blocks. Section \textcircled{1} corresponds to the attention block, and Section \textcircled{2} corresponds to the MLP block. The change of data dimension is illustrated.}
    \vspace{-30pt}
    \label{fig:figure1}
\end{figure}

LLM models~\cite{ref34,ref35,ref36,ref37} typically include the token phase as illustrated in Figure~\ref{fig:figure1}, where both the attention block and the MLP block consist of multiple kernel operators. The attention block is primarily responsible for capturing inter-token dependencies. To enhance model capability, the sequence length is progressively increased. However, due to the presence of the nonlinear ‘Softmax’ operator, the computational complexity of attention—given by $\mathcal{O}(2N_t N d + N_t N)$—grows rapidly with the sequence length, making it a major performance bottleneck in current Transformer-based systems. $N_{t}$ denotes the number of total tokens (KV Cache), $d$ is the split hidden size per head and $N$ is the current token’s sequence length (query). On the other hand, the MLP block is a feedforward network that performs complex mappings from input to output tensors through a combination of nonlinear activations and linear transformations. As these blocks dominate the parameter count in Transformer architectures, maximizing the computational efficiency of this component is essential.

In this work, we primarily consider the decoder-only Transformer executed on general-purpose AI accelerator architectures equipped with a three-level cache hierarchy (L1, L2, and L3). The cache capacities increase progressively from L1 to L3 (i.e., $L1 < L2 < L3$), while the corresponding access speeds decrease accordingly (i.e., $L1 > L2 > L3$).  As illustrated in Figure~\ref{fig:figure1}, during the LLM computation, tokens are processed sequentially during the decoder-only LLM inference. The input data, the model weights, and the intermediate results are typically stored in the L3. Tokens are represented as a tensor of shape $\mathbb{R}^{B \times N \times D}$, where $B$ denotes the batch size, and $D$ denotes the hidden size. The hidden size $D$ can be further decomposed as $D = H \times d$, where $H$ is the number of attention heads. It is important to note that during the token phase, both $B$ and $N$ are always equal to 1, indicating that only a single token is processed at a time per sequence. The computational complexity comes to $\mathcal{O}(2N_t d + N_t)$. The dataflow diagram in Figure~\ref{fig:figure1} illustrates how a single token tensor propagates through the attention and MLP blocks during the token phase.

\subsection{Standard Attention Implement}
\label{sec:section2.2}
The standard implementation of the attention block is illustrated in Section \textcircled{1} of Figure~\ref{fig:figure1}. The process begins by applying a weight matrix of shape $\mathbb{R}^{D \times D}$ to the input token of shape $\mathbb{R}^{B \times N \times D}$, resulting in the generation of three key input tensors: $Q$, $K$, and $V$, each with shape $\mathbb{R}^{B \times N \times D}$. As shown in the standard attention block flow of the TPS architecture in Figure~\ref{fig:figure1}, the \texttt{RoPE} operator requires sinusoidal tensors $\sin$ and $\cos$ of shape $\mathbb{R}^{H \times (d/2)}$. Additionally, the \texttt{Concat} operator depends on the KV Cache tensors $K_{\text{cache}}, V_{\text{cache}} \in \mathbb{R}^{B \times N \times D}$.

The computational process described above is illustrated in the flow diagram of Figure~\ref{fig:figure1} and Algorithm~\ref{alg:standard_attention}. In Step 1 and Step 2, the tensors $Q$ and $K$, along with the sinusoidal tensors $\sin$ and $\cos$, are loaded from the L3 into the L1 cache. After that, a \texttt{RoPE} is applied to $Q$ and $K$, producing $Q_{\text{rope}}$ and $K_{\text{rope}}$. Then, $Q_{\text{rope}}$, $K_{\text{rope}}$, and $V$ are loaded from the L3 into the L1, where \texttt{Transpose} operators are applied to rearrange the $H$ and $N$ dimensions, and the resulting tensors $Q_{\text{trans}}$, $K_{\text{trans}}$, and $V_{\text{trans}}$ are stored back into the L3.

Step 3 involves loading the $K_{\text{trans}}$, $V_{\text{trans}}$, and the cached tensors $K_{\text{cache}}$ and $V_{\text{cache}}$ from the L3 into the L1 to perform the KV Cache update. This is achieved by concatenating tensors along specific dimensions to form the updated $K_{\text{cache}}$ and $V_{\text{cache}}$. In Steps 4--6, $Q_{\text{trans}}$ and $K_{\text{cache}}$ are loaded from the L3 into the L1 to compute the \texttt{QKT} matrix multiplications, producing the intermediate result $S$. Next, $S$ undergoes the \texttt{Softmax} operation, producing the result $P$. Then, $P$ and $V_{\text{cache}}$ are used for the final \texttt{PV} matrix multiplication, resulting in the output tensor $O$. The inputs and outputs of multiple kernels frequently shuttle between the L1 and the L3.

\begin{algorithm}[tb]
   \caption{Standard Attention Implementation}
   \label{alg:standard_attention}
\begin{algorithmic}
   \REQUIRE Tensors $Q, K, V \in \mathbb{R}^{B \times N \times H \times d}$, $sin, cos \in \mathbb{R}^{H \times (d/2)}$, $K_{\text{cache}}, V_{\text{cache}} \in \mathbb{R}^{B \times H \times N_p \times d}$
   \STATE \textbf{Step 1:} Read $Q, K, sin, cos$ from L3$\rightarrow$L1, compute RoPE, write $Q_{\text{rope}}, K_{\text{rope}}$ from L3$\rightarrow$L1
   \STATE \textbf{Step 2:} Read $Q_{\text{rope}}, K_{\text{rope}}, V$ from L3$\rightarrow$L1, compute Transpose, write $Q_{\text{trans}}, K_{\text{trans}}, V_{\text{trans}}$ from L1$\rightarrow$L3
   \STATE \textbf{Step 3:} Read $K_{\text{trans}}, V_{\text{trans}}, K_{\text{cache}}, V_{\text{cache}}$ from L3$\rightarrow$L1, compute Concat, write updated $K_{\text{cache}}, V_{\text{cache}}$ from L1$\rightarrow$L3
   \STATE \textbf{Step 4:} Read $Q_{\text{trans}}, K_{\text{cache}}$ from L3$\rightarrow$L1, compute $QK^T$, write $S$ from L1$\rightarrow$L3
   \STATE \textbf{Step 5:} Read $S$ from L3$\rightarrow$L1, compute Softmax, write $P$ from L1$\rightarrow$L3
   \STATE \textbf{Step 6:} Read $P, V_{\text{cache}}$ from L3$\rightarrow$L1, compute $PV$, write $O$ from L1$\rightarrow$L3
   \STATE \textbf{Return:} $O \in \mathbb{R}^{B \times H \times N \times d}$
\end{algorithmic}
\end{algorithm}

The low efficiency of KV Cache utilization is the main challenge of attention implementation: as memory reorganization operators, the implementation of KV Cache \texttt{Concat} relies on substantial memory usage~\cite{ref48,ref49,ref50,ref51,ref52,ref53}. It is essential to continuously improve the memory efficiency of KV Cache. Moreover, on AI accelerator devices, the limited memory bandwidth poses an additional challenge—particularly in the later stages of token generation, where the growing size of the KV Cache leads to increasing bandwidth pressure and potential bottlenecks. In addition, the participation of the CPU introduces synchronization operations on the AI accelerators, making it difficult to perform fine-grained scheduling of operators. Consequently, ensuring high model performance on AI accelerators requires not only improving memory utilization but also fine-grained optimization, making the optimization of KV Cache a significant challenge. We will optimize KV Cache management in this article.

\subsection{Standard MLP Implement}
The standard implementation of the MLP consists of five steps, as shown in Section \textcircled{2} of Figure~\ref{fig:figure1}: 
(1) \textbf{RMSNorm}, which improves the gradient distribution during training and accelerates convergence~\cite{ref41}; 
(2) \textbf{Gate projection}, which performs feature extraction on the input tensor for subsequent nonlinear transformations; 
(3) \textbf{Up projection}, which expands the dimensionality of the input tensor, providing more information for elementwise multiplication; 
(4) \textbf{SiLU activation}~\cite{ref42}, which applies nonlinear transformation to the feature tensor after gate projection; and 
(5) \textbf{Down projection}, which compresses the features expanded from the hidden layer to the middle layer back to the hidden layer size for the next computation.

The computation is defined as:
\begin{equation}
O = W_{\text{down}} \left( \text{SiLU}(W_{\text{gate}} h_t) * (W_{\text{up}} h_t) \right)
\label{eq:mlp_main}
\end{equation}

Where $h_t \in \mathbb{R}^{N \times D}$ is the normalized input tensor by the \texttt{RMSNorm} function; $W_{\text{gate}}, W_{\text{up}} \in \mathbb{R}^{D \times K}$ and $W_{\text{down}} \in \mathbb{R}^{K \times D}$ are the weight matrices for gate, up, and down projections respectively; \texttt{SiLU} is the activation function; and $O$ is the output matrix after two-layer computation. The details are shown in Algorithm~\ref{alg:standard_mlp}. Where $h_t^g, h_t^u \in \mathbb{R}^{N \times K}$ are the feature tensors after gate and up projections respectively; $h_t^s \in \mathbb{R}^{N \times K}$ is the feature tensor after nonlinear transformation by the \texttt{SiLU} activation function; $h_t' \in \mathbb{R}^{N \times K}$ is the result of elementwise multiplication between $h_t^u$ and $h_t^s$; and $O \in \mathbb{R}^{N \times D}$ is the final output. 

MLP's core computations are dominated by matrix multiplications. However, since $N = 1$, the computational parallelism decreases dramatically, leading to underutilization of matrix multiplication acceleration units. We use dataflow to mitigate this problem which can be strategically organized to pipeline memory and computing operations, thereby effectively hiding the memory latency and improving the overall throughput.

\begin{algorithm}[tb]
   \caption{Standard MLP Implementation}
   \label{alg:standard_mlp}
\begin{algorithmic}
   \REQUIRE Tensor input $h_t \in \mathbb{R}^{N \times D}$; weights $W_{\text{gate}}, W_{\text{up}} \in \mathbb{R}^{D \times K}$, $W_{\text{down}} \in \mathbb{R}^{K \times D}$
   \STATE \textbf{Step 1:} Read $h_t$ and $W_{\text{gate}}$ from L3$\rightarrow$L1, compute gate projection $W_{\text{gate}} h_t$, write $h_t^g \in \mathbb{R}^{N \times K}$ from L1$\rightarrow$L3
   \STATE \textbf{Step 2:} Read $h_t^g$ from L3$\rightarrow$L1, compute SiLU activation $\text{SiLU}(h_t^g)$, write $h_t^s$ from L1$\rightarrow$L3
   \STATE \textbf{Step 3:} Read $h_t$ and $W_{\text{up}}$ from L3$\rightarrow$L1, compute up projection $W_{\text{up}} h_t$, write $h_t^u \in \mathbb{R}^{N \times K}$ from L1$\rightarrow$L3
   \STATE \textbf{Step 4:} Read $h_t^s$ and $h_t^u$ from L3$\rightarrow$L1, compute element-wise multiplication $h_t' = h_t^u \odot h_t^s$, write $h_t' \in \mathbb{R}^{N \times K}$ from L1$\rightarrow$L3
   \STATE \textbf{Step 5:} Read $h_t'$ and $W_{\text{down}}$ from L3$\rightarrow$L1, compute down projection $W_{\text{down}} h_t'$, write $O \in \mathbb{R}^{N \times D}$ from L1$\rightarrow$L3
   \STATE \textbf{Return:} $O \in \mathbb{R}^{N \times D}$
\end{algorithmic}
\end{algorithm}

\subsection{Operator Type and OI}
As illustrated in the flow diagram of Figure~\ref{fig:figure1} and the corresponding algorithmic steps, the operators can be broadly categorized into two groups: (1) \textbf{Computational operators}, which include \texttt{RoPE}, \texttt{Matmul}, \texttt{Softmax}, \texttt{RMSNorm}, \texttt{Mul}, and \texttt{siLu}. These operators correspond to arithmetic computations. We need to maximize the degree of parallelism of these operators to improve computing efficiency; (2) \textbf{Memory reorganization operators}, represented by \texttt{Concat}, which involves reading both new and cached data from memory, performing concatenation, and writing the result back to the memory. This includes KV Cache or quantization KV Cache operations. This operator is purely memory-bound and does not involve any arithmetic computation. We need to improve their memory usage efficiency as much as possible to improve the bandwidth utilization. Both types of operators have a significant impact on latency and power consumption during model inference~\cite{ref42,ref43,ref44}.

To quantitatively analyze performance, we study the metric of OI which is used as a proxy to quantify the maximum potential performance of an individual operator, based on a set of hardware. The OI is defined as the number of arithmetic operations divided by the number of memory accesses. A lower OI indicates limited data reuse within the kernel and a higher likelihood of encountering memory bandwidth bottlenecks. This metric serves as an indicator of the theoretical upper bound of kernel performance on the underlying accelerator architecture.

\begin{figure}[htbp]
    \centering
    \includegraphics[width=0.5\textwidth]{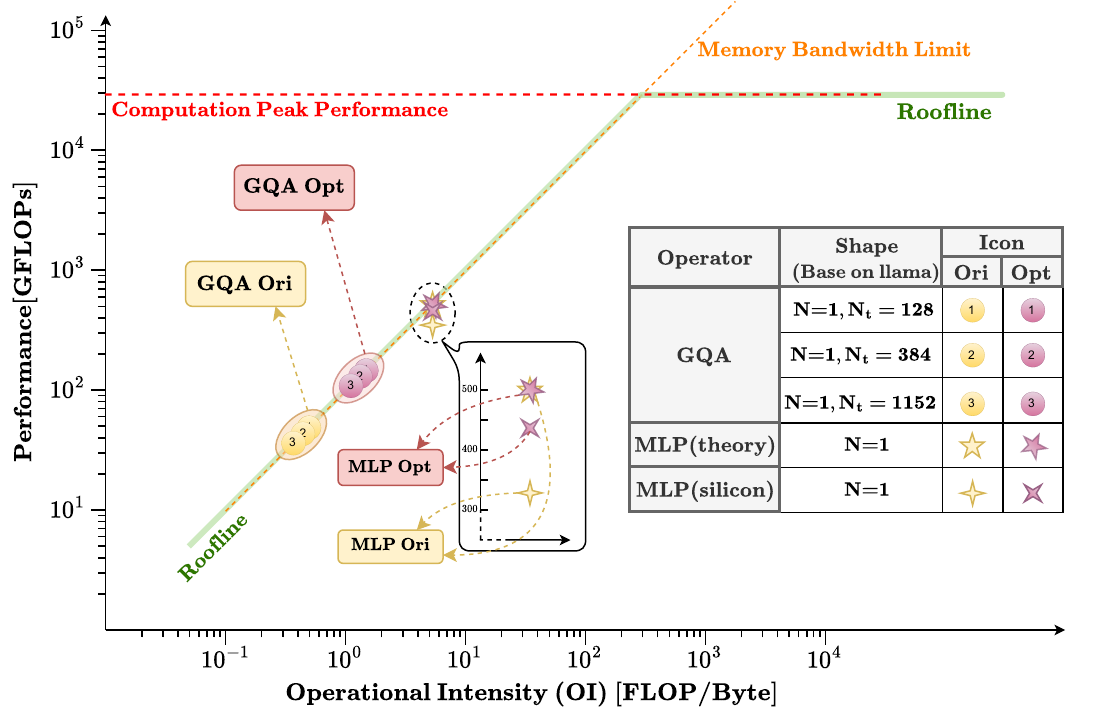}
    \caption{\small Roofline analysis of the attention and MLP blocks for some specific $N_{\text{t}}$, where the attention block incorporates the GQA (grouped query attention) optimization. \texttt{Ori} refers to the original implementation before FastTPS optimization, and \texttt{Opt} indicates the optimized version following the application of FastTPS. For the attention block, the analysis covers scenarios where the number of tokens gradually increases. For the MLP block, both theoretical performance bounds and empirical runtime results are presented. The computation peak performance of AMD Ryzen AI 300 series NPU ~\cite{ref31,ref32,ref33} is used here ($2.8^4$ GFLOPs).  }
    \vspace{-15pt}
    \label{fig:figure2}
\end{figure}

\section{FASTTPS IMPLEMENTATION}

To optimize the token phase, this paper proposes \textbf{FastTPS}, which introduces three optimization strategies targeting the attention and MLP blocks of TPS: (1) \textbf{GKVC}: This approach ensures that the KV Cache process is entirely executed on AI accelerators, maintaining computational continuity on the same hardware device. It also eliminates the need for memory reorganization operations, thereby reducing memory read/write operations and optimizing the KV Cache pipeline; (2) \textbf{TPSFLAT}: Inspired by the FLAT concept and leveraging the improved operator continuity enabled by GKVC on AI accelerators, this method designs a dataflow tailored to TPS characteristics to enable further operator fusion; (3) \textbf{Fusion MLP}: A strategy is adopted to enhance computational efficiency within the MLP block by means of decomposing and fusing operators as well as optimizing pipeline scheduling. All the above strategies preserve computational accuracy. The following sections provide a detailed description of the implementation of these optimization strategies.

\subsection{Attention Block Implementation}
Figure~\ref{fig:Figure3} illustrates the optimized processing pipeline of the attention block under KV Cache and TPSFLAT. Corresponding to Figure~\ref{fig:Figure3}, Algorithm~\ref{alg:optimized_attention} demonstrates that the proposed approach significantly reduces the number of data transfers and enhances computational continuity. A detailed explanation of the implementation is as Algorithm~\ref{alg:optimized_attention}.

\begin{algorithm}[tb]
   \caption{GKVC \& TPSFLAT}
   \label{alg:optimized_attention}
\begin{algorithmic}[1]
   \REQUIRE Tensors $Q, K, V \in \mathbb{R}^{B \times N \times H \times d}$, $sin, cos \in \mathbb{R}^{H \times (d/2)}$, KV Cache $ \in \mathbb{R}^{B \times H \times N_r \times d}$; In TPS, $B=1$, $N=1$

   \STATE \hspace{0pt} \textbf{Do} GKVC for $V$ when writing it from L1$\rightarrow$L3 after activation; get $V_{\text{cache}} \in \mathbb{R}^{B \times H \times N_t \times d}$ in L3

   \STATE \hspace{0pt} \textbf{for} $H_i$ in range $(0, H, H_b)$ \textbf{do} \COMMENT{Outer loop – L3 tile}
   \begin{ALC@g}
       \STATE \textbf{Read} $Q_b, K_b \in \mathbb{R}^{B \times N \times H_b \times d}$ and $V_{\text{cache}_b} \in \mathbb{R}^{B \times H_b \times N_t \times d}$ from L3$\rightarrow$L2
       \STATE \textbf{Read} $sin_b, cos_b \in \mathbb{R}^{H_b \times (d/2)}$ from L3$\rightarrow$L2

       \STATE \textbf{for} $h_i$ in range $(H_i, H_i + H_b, 1)$ \textbf{do} \COMMENT{Inner loop – L2 tile}
       \begin{ALC@g}
           \STATE \textbf{Read} $Q_i, K_i \in \mathbb{R}^{B \times N \times 1 \times d}$ from L2$\rightarrow$L1 as one continuous tensor $QK_i \in \mathbb{R}^{2 \times B \times N \times 1 \times d}$
           \STATE \textbf{Read} $sin_i, cos_i \in \mathbb{R}^{1 \times (d/2)}$ from L2$\rightarrow$L1
           \STATE \textbf{Compute} RoPE once, get $QK_{i_{\text{rope}}} \in \mathbb{R}^{2 \times B \times N \times 1 \times d}$ in L1, same as two continuous tensors: $Q_{i_{\text{rope}}}, K_{i_{\text{rope}}} \in \mathbb{R}^{B \times 1 \times N \times d}$ . \text{In TPS, } $B = 1$ \text{ and } $N = 1$, \text{ so it can be seen as } $Q_{i_{\text{rope}}}, K_{i_{\text{rope}}} \in \mathbb{R}^{B \times (H=1) \times N \times d}$     
           \STATE \textbf{Write} $K_{i_{\text{rope}}}$ from L1$\rightarrow$L3 with GKVC to update $K_{\text{cache}} \in \mathbb{R}^{B \times H \times N_t \times d}$ in L3
           \STATE \textbf{Read} $K_{\text{cache}_i} \in \mathbb{R}^{B \times 1 \times N_t \times d}$ from L3$\rightarrow$L1. \textbf{Compute} $QK^T(Q_{i_{\text{rope}}}, K_{\text{cache}_i})$, get $S_i \in \mathbb{R}^{B \times 1 \times N \times N_t}$ in L1
           \STATE \textbf{Compute} $\text{Softmax}(S_i)$, get $P_i \in \mathbb{R}^{B \times 1 \times N \times N_t}$ in L1
           \STATE \textbf{Read} $V_{\text{cache}_i} \in \mathbb{R}^{B \times 1 \times N_t \times d}$ from L2$\rightarrow$L1. \textbf{Compute} $PV(P_i, V_{\text{cache}_i})$, get $O_i \in \mathbb{R}^{B \times 1 \times N \times d}$ in L1
       \end{ALC@g}

       \STATE \textbf{Get} $O_b \in \mathbb{R}^{B \times H_b \times N \times d}$ in L2
   \end{ALC@g}

   \STATE \hspace{0pt} \textbf{Return:} $O \in \mathbb{R}^{B \times H \times N \times d}$ in L3
\end{algorithmic}
\end{algorithm}

\subsubsection{Global KV Cache Management}

Compared with the standard KV Cache implementations discussed in Section~\ref{sec:section2.2}, GKVC ensures that TPS inference is consistently executed on AI accelerators while eliminating the reliance on the \texttt{Concat} operator responsible for KV Cache memory reorganization. The integration of this optimization into the TPS pipeline is illustrated in Figure~\ref{fig:Figure3}.

To support this scheme, GKVC requires a pre-allocated memory region in the L3 cache for KV Cache storage, which is also known as \textit{static KV Cache}~\cite{ref40}. Given that the shape of a single $K$ or $V$ tensor is $\mathbb{R}^{B \times N \times H \times d}$, and the $N$ dimension increases progressively during TPS inference, the maximum sequence length that can be stored in the pre-allocated memory is predetermined when $B$, $H$, and $d$ are fixed to 1.

The existing KV Cache tensors are of shape $\mathbb{R}^{B \times H \times N_p \times d}$, and the cache update proceeds by directly writing the new KV Cache data into the corresponding positions in the remaining cache space. This process can be executed either by the AI accelerator or by software.

Specifically, for each iteration over the $B \times H$ dimension, denoted as $h_i = [0 : B \times H)$, a vector of length $d$ is written sequentially after a jump of:
\begin{equation}
\text{Step} = h \cdot (N_p + N_r) \cdot d + N_p \cdot d, \quad h = 0, 1, 2, \ldots, B \cdot H - 1
\label{eq:address_jump}
\end{equation}
Once the $B \times H$ iterations are complete, the KV Cache update is finalized. The sequence length in the cache is incremented to $N_t$, and $N_p$ is updated to $N_t$, while $N_r$ is decremented by 1 accordingly.

GKVC uses an address-jumping strategy to write into pre-allocated memory processed by AI accelerator (or software), thereby eliminating the data read/write and concatenation executed by CPU. This scheme enables the entire inference pipeline to remain on the AI accelerator, which allows subsequent operator fusion optimization to theoretically support the fusion of more operators.

\subsubsection{TPSFLAT}

Based on the FLAT~\cite{ref30} scheme, this paper proposes a specialized dataflow strategy tailored to the characteristics of the attention block with \texttt{RoPE} in TPS, termed \textbf{TPSFLAT}. As introduced in FLAT, \texttt{Softmax} performs normalization along a specific dimension, which imposes a minimum granularity of computation—normalization is performed whenever the input reaches the minimal computation granularity, and the output is then passed to the next operator. This enables the decomposition of the computation task into multiple parallel processes with independent data flows, based on the normalization dimension of \texttt{Softmax}. In TPS, both the $B$ and $N$ dimensions are set to 1. Therefore, the computation task can be decomposed along the $H$ dimension, resulting in a simplified tiling strategy.

\begin{figure}[htbp]
    \centering
    \includegraphics[width=0.48\textwidth]{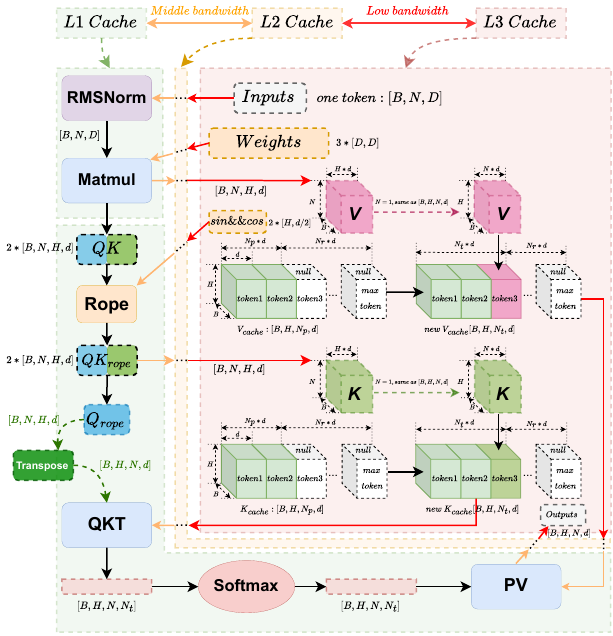}
    \vspace{-10pt}
    \caption{\small Structural overview of attention blocks after GKVC and TPSFLAT optimization. The red, orange, and green areas correspond to the L3, L2, and L1 cache respectively. In GKVC, the KV vectors computed in the L1 cache are directly written to the pre-allocated memory locations in the L3 cache via the L2 cache, eliminating the need for a \texttt{Concat} operator. This strategy ensures memory without fragmentation. TPSFLAT achieves a continuous fusion of operations—including \texttt{RoPE,} \texttt{QKT,} \texttt{Softmax,} and \texttt{PV}—within the L1 cache, the significantly reduces data transfer and memory access overhead between operations.}
    \vspace{-1pt}
    \label{fig:Figure3}
\end{figure}

The inference pipeline optimized by the GKVC strategy proposed in this paper eliminates the need for the \texttt{Concat} operator while maintaining the entire inference pipeline on the AI processor. With GKVC, we fuse the \texttt{RoPE} operator before KV Cache and inherit the fusion of L/A operators from FLAT (corresponding to \texttt{QKT}, \texttt{Softmax}, and \texttt{PV} in this paper). As illustrated in Algorithm~\ref{alg:optimized_attention}, Steps 6 and 7, considering that $Q$ and $K$ share the same \texttt{RoPE} logic, our scheme applies the \texttt{RoPE} operator jointly to $Q$ and $K$, allowing the sinusoidal tensors to be shared. This reduces operator calls. As described in Algorithm~\ref{alg:standard_attention}, the native inference pipeline executes operators sequentially. After applying TPSFLAT, only the necessary KV tensors are written back to L3 for KV Cache, while all other intermediate tensors are retained in L1 for subsequent computations. The operator fusion in TPSFLAT is demonstrated in both Figure~\ref{fig:Figure3} and Algorithm~\ref{alg:optimized_attention}, which illustrate the data flow across the L1, L2, and L3 caches. This significantly reduces the number of transfers between the L1 and L3 caches, saving transmission time and ultimately improving inference speed.

\subsubsection{Tiling Strategy and Execution Granularity}
TPSFLAT inherits the characteristics of FLAT in terms of data dependencies and avoiding excessive on-chip tensor sizes. To enable operator fusion, this approach also requires partitioning the input data into smaller blocks along specific dimensions. These blocks are then processed through nested loops to complete the full computation workload. This partitioning strategy is referred to as \textbf{tiling}. The tiling configuration varies across different AI accelerators.

In this paper, we take the AMD Ryzen AI 300 series NPU~\cite{ref31,ref32,ref33} as an example, which is detailed in Algorithm~\ref{alg:optimized_attention}. It is known that before fusion was adopted, each kernel’s execution involves input and output data accesses to the L3 cache. Consequently, each L1 access can only support the computation of a single kernel.

In contrast, after fusion, four kernels can be executed consecutively within the L1 cache, significantly reducing memory access frequency and improving the OI. The detailed computation processes after tiling are described in Appendix~\ref{appendix:A1}, and the usages in different cache levels are analyzed in Appendix~\ref{appendix:A2}.

\subsection{Fusion MLP Implementation}
\label{sec:section4.2}

\begin{figure}[htbp]
    \centering
    \includegraphics[width=0.5\textwidth]{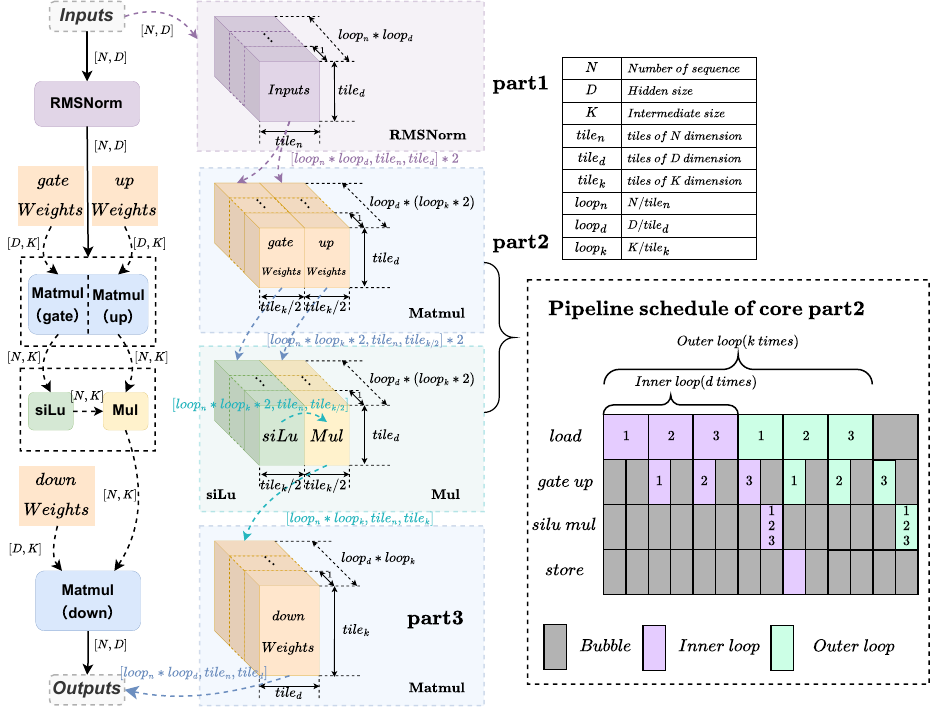}
    \caption{\small The Fusion MLP workflow, each part represents the entire execution process of data exchange between the L1 cache and the L3 cache. The part1 is the \texttt{RMSNorm} operation applied to normalize the input tensor. The part2 is the primary part of MLP which describes the methods of interlaced \texttt{gate}-\texttt{up} weights. The pipeline schedule in part2 demonstrates improvement by introducing interlaced weights which eliminates the time-span due to OPs' sequential execution. The part3 is the down projection operation served to reduce the dimension enriched by part2 for subsequent calculations.}
    \vspace{-1pt}
    \label{fig:figure4}
\end{figure}

MLP is a type of feedforward network~\cite{ref39, ref56} that processes the complex mapping relationship from input tensors to output tensors through the combination of nonlinear functions and linear transformations. The MLP~\cite{ref35, ref57} includes five operations: gate projection, up projection, and down projection, as well as \texttt{siLu} operation and \texttt{elewMul} operation. The conventional approach is to fuse gate projection (compute-bound) and \texttt{siLu} operators (memory-bound) together in software graph fusion passes to reduce I/O overhead and balance computation time with I/O costs. Consequently, this produces four operators in MLP: \texttt{gate\_siLu}, up projection, \texttt{elewMul}, and down projection. To further integrate operators and reduce I/O overhead, we propose a novel function \textbf{Fusion MLP} including two stages: (1) fuse the gate and the up projections into a single large \texttt{Matmul}; (2) fuse the combined \texttt{Matmul} with the \texttt{siLu} and \texttt{elewMul} operators. This method reduces the number of I/Os from 7 to 2 and enables a fine-grain pipeline which can enhance computational efficiency, maximize resource utilization, and bring the computing performance closer to the theoretical peak, as illustrated in Figure~\ref{fig:figure2}. It contains three key optimization points:

\begin{enumerate}

    \item \textbf{Interlaced layout of gate-up weights:} Assuming the minimum data unit required for computation on the L1 cache is a tile block with shape $[\text{tile}_n, \text{tile}_d] \times [\text{tile}_d, \text{tile}_k]$, we split the weight matrices $W_{\text{gate}}, W_{\text{up}}$ along the retained dimension $K$ into half tile blocks with shape $[\text{tile}_d, \text{tile}_k/2]$, then rearrange these blocks in the order of block $W_{\text{gate}}$ and block $W_{\text{up}}$ into a new tile. This method constructs a larger \texttt{Matmul} (blue block in part 2 of Figure~\ref{fig:figure4}) while providing necessary conditions for subsequent operator fusion and pipeline scheduling.

    \item \textbf{Operator fusion:} Based on the interlaced layout of gate-up weights, the \texttt{siLu} and \texttt{elewMul} are fused with the larger \texttt{Matmul} to reduce memory access frequency between the L1, L2, and L3 caches, thereby improving OI. As described in Figure~\ref{fig:figure4} part 2, the \texttt{siLu} operation is executed after the gate part of the larger \texttt{Matmul}. Then, \texttt{Mul} is executed by multiplying the matrix generated by the \texttt{siLu} operation and the up projection. This fused operator is renamed to \texttt{siLu\_Mul}. It is worth mentioning that other elementwise operators besides \texttt{siLu} and \texttt{elewMul} can also be optimized in the same way.
    
    \item \textbf{Pipeline scheduling:} Pipeline scheduling is employed within the larger \texttt{Matmul} to increase data usage efficiency. We identify weight loading as the primary performance bottleneck in the token phase of MLPs. To address this, we maximize weight reuse and overlap computation with I/O operations to mask latency.    
    
\end{enumerate}

\begin{algorithm}[tb]
   \caption{MLP Implementation}
   \label{alg:mlp_implementation}
\begin{algorithmic}[1]
   \REQUIRE Tensors in L3: $h_t \in \mathbb{R}^{N \times D}$, $W_{\text{gu}} \in \mathbb{R}^{D \times 2K}$, $W_{\text{down}} \in \mathbb{R}^{K \times D}$, $h_t^{\text{sm}} \in \mathbb{R}^{N \times K}$, $O \in \mathbb{R}^{N \times D}$; tile dimensions $\text{tile}_n \in \{1, N\}$, $\text{tile}_k \in \{1, 2K\}$, $\text{tile}_d \in \{1, D\}$

   \STATE \textbf{Do} \texttt{RMSNorm} for the input to get hidden states $h_t$
   
   \STATE \textbf{for} $n = 1$ to $\lceil N / (x \cdot \text{tile}_n) \rceil$ $k = 1$ to $\lceil 2K / (y \cdot \text{tile}_k) \rceil$
       \begin{ALC@g}
           \STATE \textbf{for} $d = 1$ to $\lceil D / (z \cdot \text{tile}_d) \rceil$
           \begin{ALC@g}
               \STATE \textbf{Load} data $h_t^{nkd}$, $W_{\text{gu}}^{nkd}$ from L3$\rightarrow$L2
               
               \STATE \textbf{for} $i = 1$ to $x \cdot y$
               \begin{ALC@g}
                   \STATE \textbf{for all} $z$
                   \begin{ALC@g}
                       \STATE \textbf{Load} $h_t^i \in \mathbb{R}^{\text{tile}_n \times \text{tile}_d}$ and $W_{\text{gu}}^i \in \mathbb{R}^{\text{tile}_d \times \text{tile}_k}$ from L2$\rightarrow$L1
                       \STATE \textbf{Compute} fused gate-up projection: $W_{\text{gu}}^i h_t^i$
                       \STATE \textbf{Accumulate} to get $h_t^{\text{gu}_i} \in \mathbb{R}^{\text{tile}_n \times \text{tile}_k}$
                   \end{ALC@g}
               \end{ALC@g}
               
               \STATE \textbf{Accumulate} to get $h_t^{\text{gu}_{nk}} \in \mathbb{R}^{x \cdot \text{tile}_n \times y \cdot \text{tile}_k}$
               \STATE \textbf{Compute} $\texttt{siLu\_Mul}(h_t^{\text{gu}_{nk}})$ to get $h_t^{\text{sm}_{nk}} \in \mathbb{R}^{x \cdot \text{tile}_n \times y \cdot (\text{tile}_k / 2)}$
               \STATE \textbf{Store} $h_t^{\text{sm}_{nk}}$ from L2$\rightarrow$L3
           \end{ALC@g}
       \end{ALC@g}

   \STATE \textbf{Load} $h_t^{\text{sm}}$ and $W_{\text{down}}$ from L3$\rightarrow$L1
   \STATE \textbf{Compute} $W_{\text{down}} h_t^{\text{sm}}$ to get output $O$
   \STATE \textbf{Store}  $O$ from L1$\rightarrow$L3
\end{algorithmic}
\end{algorithm}

The process details are shown in Algorithm~\ref{alg:mlp_implementation} and Figure~\ref{fig:figure4}. In Algorithm~\ref{alg:mlp_implementation}, the input tensor is $h_t \in \mathbb{R}^{N \times D}$, $W_{\text{gu}} \in \mathbb{R}^{D \times 2K}$, and $W_{\text{down}} \in \mathbb{R}^{K \times D}$, where $h_t$ is the input tensor normalized by \texttt{RMSNorm}, and $W_{\text{gu}}$ is the new weight matrix formed by interlacing the gate and up weight matrices. The dimensions $\text{tile}_n \in \{1, N\}$, $\text{tile}_k \in \{1, 2K\}$, and $\text{tile}_d \in \{1, D\}$ represent the minimum tile block dimensions required for a single core matrix multiplication in the L1 cache. The quantities $x$, $y$, and $z$ denote the number of $\text{tile}_n$, $\text{tile}_k$, and $\text{tile}_d$ tiles that can be loaded or stored between the L3 cache and L2 cache, respectively. These parameters can be adjusted dynamically based on the data transfer bandwidth, computational bandwidth, and the L2 cache capacity, thereby achieving the optimal partition strategy. See Appendix~\ref{appendix:A3} for more descriptions.

\section{EXPERIMENTS}
In this section, we evaluate the inference latency speedup achieved through optimizations on two key computational components of Transformer architectures: (1) the attention block (Section~\ref{sec:section5.1}) and (2) the MLP block (Section~\ref{sec:section5.2}), as implemented on the AMD Ryzen AI 300 series NPU~\cite{ref31, ref32, ref33} of several state-of-the-art LLMs, including ChatGLM3-6B (ChatGLM)~\cite{ref34}, Llama2-7B (Llama2)~\cite{ref35}, Llama3-8B-instruct (Llama3)~\cite{ref36}, Llama3.2-1B (Llama3.2)~\cite{ref36} and Phi3-mini-4k-instruct (Phi3)~\cite{ref37}, using standard model implementations from the HuggingFace. We further examine the performance of these LLMs on this hardware platform (Section~\ref{sec:section5.3}). In these tests, the CPU overhead wasn’t taken into consideration. Finally, we analyze the OI and numerical accuracy characteristics -- the primary optimization targets of our proposed method (Section~\ref{sec:section5.4}).

\subsection{Faster Attention Block}
\label{sec:section5.1}
The models and their configurations presented in Table~\ref{table-model-config} were evaluated. Among these, ChatGLM and Llama3 have been specifically optimized using GQA~\cite{ref58}, a technique designed to mitigate the memory bandwidth overhead associated with KV Cache loading. In this implementation, GQA operates by broadcasting each query head’s computation to its corresponding group of key and value heads, thereby reducing the memory access requirements while maintaining computational accuracy.

\begin{table}[t]
\vskip -0.05in
\caption{Model configurations used in our test.}
\label{table-model-config}
\vskip 0.05in
\begin{center}
\begin{scriptsize}
\begin{tabular}{lccccc}
\toprule
Model & L3 & GLM & L2 & P3 & L3.2 \\
\midrule
Q. Heads     & 32    & 32    & 32    & 32   & 32 \\
KV Heads     & 8     & 2     & 32    & 32   & 32 \\
Head Dim     & 128   & 128   & 128   & 96   & 64 \\
Max $N$      & 2k/4k & 2k/4k & 2k    & 2k   & 2k \\
MLP $K$      & 14336 & 13696 & 11008 & 8192 & 8192 \\
\bottomrule
\end{tabular}
\end{scriptsize}
\end{center}
\vskip -0.1in
\end{table}

\begin{figure}[htbp]
    \centering
    \includegraphics[width=0.48\textwidth]{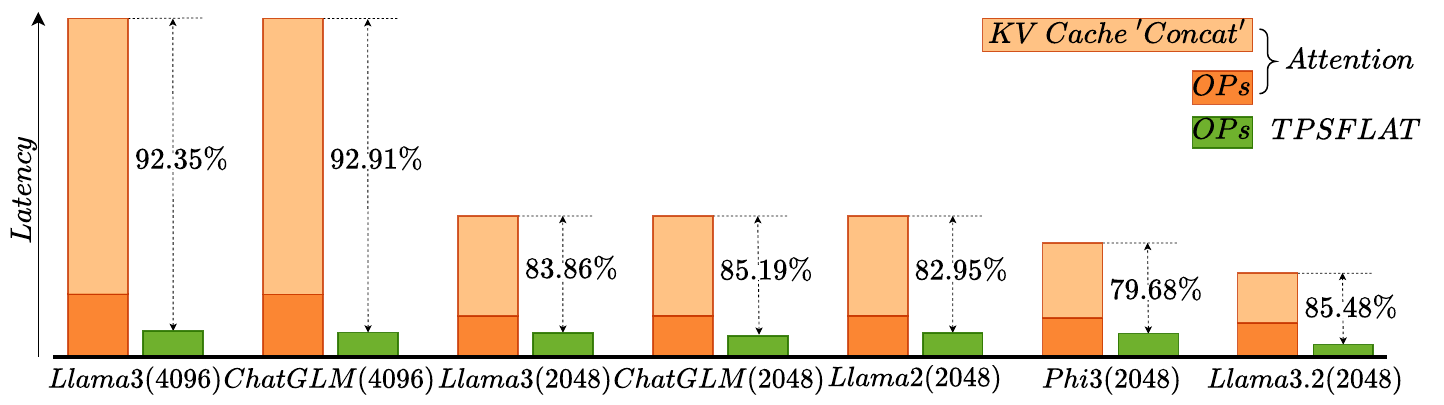}
    \vspace{-15pt}
    \caption{\small Latency of different models in Table~\ref{table-model-config} of attention and TPSFLAT. The orange bar and green bar are the latency of attention and TPSFLAT respectively. Besides, attention shows the latency ratio of KV Cache `Concat` (light orange) and other operations (dark orange). The number above the green bar represents the reduction in latency achieved by TPSFLAT.}
    \vspace{-1pt}
    \label{fig:Figure5}
\end{figure}

Figure~\ref{fig:Figure5} presents the latency performance of various transformer models, with numerical annotations above the green bars indicating the percentage reduction in TPSFLAT latency relative to the attention baseline. The attention mechanism involves KV Cache concatenation (processed by the CPU) and subsequent computations on the AI accelerator which leads to low OI. Our proposed TPSFLAT optimizations eliminate these overheads---KV Cache `Concat` (66\%$\sim$82\%, light orange in Figure~\ref{fig:Figure5}) and calculation costs (18\%$\sim$34\%, dark orange in Figure~\ref{fig:Figure5}) respectively---delivering significant performance gains, as described in Section~4. The largest latency reduction is 93\% for Llama3(4096), with 80\%$\sim$92\% improvements across other models. Computational costs are reduced by 40\%$\sim$64\% across all models. Notably, Llama3, Llama2, and ChatGLM share identical architectural configurations at the same max N, differing only in their query head counts. The observed latency disparity among these models is limited to 7.2\%, demonstrating that our GQA implementation introduces negligible computational overhead.

\begin{figure}[htbp]
    \centering
    \vspace{-15pt}
    \includegraphics[width=0.4\textwidth]{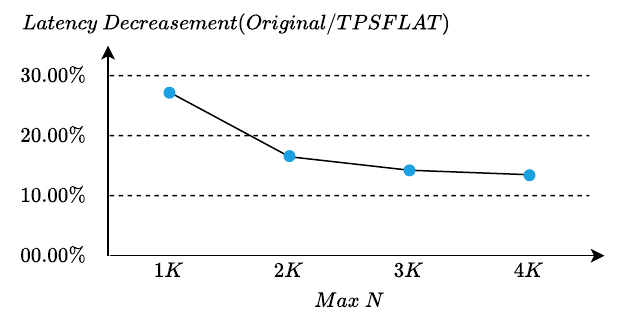}
    \vspace{-10pt}
    \caption{\small Latency reduction rate of different max sequence lengths for Llama2. }
    \vspace{-15pt}
    \label{fig:Figure6}
\end{figure}

Empirical analysis reveals a quadratic relationship between latency and sequence length ($N_t$), as evidenced by the dominant $\mathcal{O}(2N_t d)$ complexity of KV Cache `Concat` operations in attention implementations. By implementing GKVC, we eliminate the `Concat` overhead, enabling significant performance acceleration as $N$ increases. To quantify this relationship, we performed systematic benchmarking across varying sequence lengths. As illustrated in Figure~\ref{fig:Figure6}, the acceleration ratio exhibits a monotonic increase with $N_t$ (optimized 70\%$\sim$90\%) for Llama2, confirming superior performance gains for extended sequence processing. The KV Cache `Concat` overhead increases as max sequence length increases.

Conversely, the observed latency reduction exhibits a clear correlation with head dimensionality ($d$). This stems from the fact that reducing $d$ significantly decreases KV Cache concatenation overhead and the computational cost of `BMM1`, while `BMM2`, `RoPE`, and `Softmax` operations remain largely unaffected. In the original implementation, large $d$ values result in substantial KV Cache participation, whereas this contribution diminishes for smaller $d$. Notably, our TPSFLAT (GKVC) approach eliminates KV Cache participation entirely. Consequently, while acceleration initially shows a slight decline for very small $d$ values (due to other fixed costs), it subsequently increases. This non-monotonic trend aligns with empirical observations for Phi3 and Llama3.2 compared to Llama2, thereby validating our method's effectiveness in optimizing high-dimensional attention mechanisms.

\subsection{MLP Block with Better Speed}
\label{sec:section5.2}
\begin{figure}[htbp]
    \centering
    \includegraphics[width=0.4\textwidth]{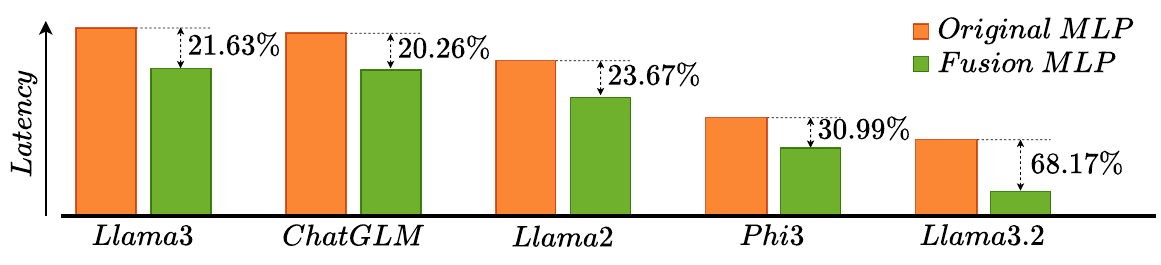}
    \vspace{-5pt}
    \caption{\small The latency of original MLP (orange bars) and Fusion MLP (green bars). The number above the green bar is the improvement ratio from Fusion MLP to original MLP. }
    \vspace{-5pt}
    \label{fig:Figure7}
\end{figure}

We conducted a comprehensive analysis of the MLP block latency across five transformer models detailed in Table~\ref{table-model-config}. As illustrated in Figure~\ref{fig:Figure7}, our experiments revealed significant latency reductions through MLP fusion: Llama3, ChatGLM, Llama2 and Phi3 demonstrated speedups exceeding 20\% compared to their unfused counterparts while Llama3.2 achieved a remarkable 68\% reduction in MLP latency. The significant performance improvements are driven by Fusion MLP’s reduction of data movement, in contrast to the original MLP’s implementation. The observed speedup ratio correlates closely with the input data proportion relative to the total read operations, approximated by the formula: $\sim \frac{K + 3D}{3KD + 4K + 4D}$ where $K$ and $D$ correspond to the description in Algorithm~\ref{alg:mlp_implementation} in Section~\ref{sec:section4.2}. This relationship quantitatively explains the varying speedup magnitudes across different model architectures.

\subsection{The Speedup of FastTPS on LLMs}
\label{sec:section5.3}
To date, we have separately evaluated the optimization effects of FastTPS on the attention block and MLP block. However, the practical significance of these optimizations within the actual LLM architecture remains to be systematically assessed. As demonstrated in Table~\ref{table-block-proportion}, the computational contributions of these blocks vary significantly across models: the attention block accounts for approximately 18$\sim$49\% of the total operations, while the MLP block constitutes 21$\sim$36\% of the total operations. These findings suggest that FastTPS would yield particularly substantial performance improvements in these model architectures.

The experimental results, summarized in Table~\ref{table-fasttps-performance}, report the measured TPS with and without FastTPS. The table also shows the FastTPS speedup and the theoretical maximum performance (calculated as the weight data divided by full memory bandwidth). Among the models tested, Phi3 achieves the highest acceleration of 6.0$\times$, attributed to its largest attention block and MLP block participation. Notably, the models achieve an average accelerator-resource utilization of 86\% (with Phi3 reaching an exceptional 93\%), demonstrating FastTPS’s effectiveness in real-world LLM deployments.

FastTPS has been integrated with AMD NPU and evaluated in comparison with other NPUs available in the industry. As presented in Tables 5-8 in Appendix \textbf{A.4}, the results indicate that during the bandwidth-limited token generation phase, the AMD NPU achieves superior performance on certain models despite its lower memory bandwidth.

\begin{table}[t]
\caption{The proportion of attention and MLP block for different models.}
\label{table-block-proportion}
\vskip -30pt
\begin{center}
\begin{small}
\begin{sc}
\begin{tabular}{lccc}
\toprule
Model & Llama3.2 & Llama2 & Phi3 \\
\midrule
Attention block (\%) & 18.28 & 42.27 & 49.35 \\
MLP block (\%)       & 36.34 & 24.31 & 20.82 \\
\bottomrule
\end{tabular}
\end{sc}
\end{small}
\end{center}
\vskip -0.1in
\end{table}

\begin{table}[t]
\caption{Latency and TPS ratios with and without FastTPS.}
\label{table-fasttps-performance}
\vskip 0.1in
\begin{center}
\begin{scriptsize}
\begin{tabular}{lccc}
\toprule
Model & L3.2 & L2 & P3 \\
\midrule
Theoretical TPS (ms)     & 37.86 & 15.56 & 18.97 \\
TPS w/o FastTPS (ms)     & 12.21 & 2.34  & 2.95  \\
TPS with FastTPS (ms)    & 32.27 & 12.41 & 17.60 \\
w/o FastTPS Ratio (\%)   & 32.24 & 15.05 & 15.58 \\
FastTPS Ratio (\%)       & 85.22 & 79.74 & 92.77 \\
Speedup                  & 2.64$\times$ & 5.30$\times$ & 5.95$\times$ \\
\bottomrule
\end{tabular}
\end{scriptsize}
\end{center}
\vskip -0.1in
\end{table}

\subsection{OI and Accuracy}
\label{sec:section5.4}
\textbf{OI Improvement |} The OI of both attention and MLP blocks remains in a low-performance region due to the token-by-token iteration inherent in the token phase. As illustrated in Figure~\ref{fig:figure2}, we present the roofline analysis for these blocks in the Llama2 model.

For the attention block, we implement several key optimizations in TPSFLAT: (1) eliminate high-overhead data movement operations in KV Cache management by redesigning the memory access pattern; (2) integrate `Softmax` and `Matmul` operations through TPSFLAT fusion, shifting the memory synchronization point from the L3 to the L1 cache; (3) combine two `RoPE` operations into a single calculation to further reduce computational overhead. These optimizations collectively minimize data I/O operations and significantly enhance OI. As shown in Figure~\ref{fig:figure2}, the yellow-scatter region (within the yellow circle) represents the OI before TPSFLAT optimization (GQA Ori.), while the red-scatter region (within the red circle) demonstrates the improved OI after optimization (GQA Opt.) by 3x.

For the MLP block, our optimization effectively reduces the computation unit wastage. Consequently, the Fusion MLP block achieves performance much closer to the theoretical maximum compared to the eager execution baseline (as shown in the dashed circle of Figure~\ref{fig:figure2}, where the purple star is much closer to the theoretical performance than the yellow star). We found the performance of MLP is restricted by the movement of weight data (vastly exceeds input) which is inevitable.

\begin{figure}[htbp]
    \centering
    \includegraphics[width=0.4\textwidth]{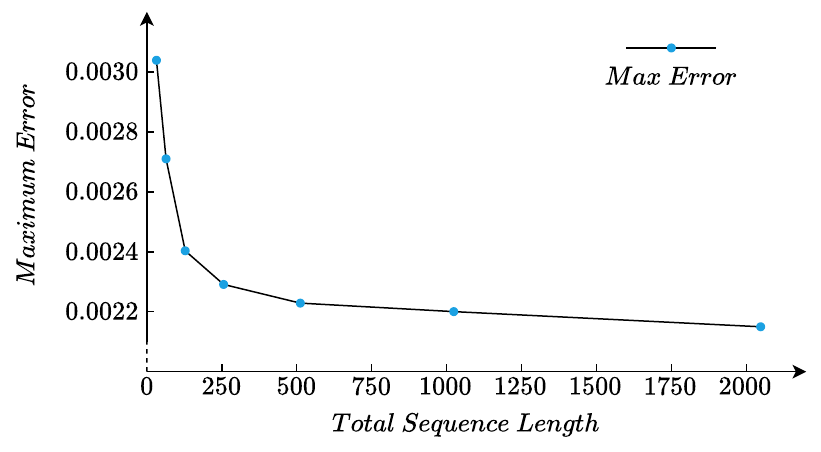}
    \vspace{-5pt}
    \caption{\small Maximum error as sequence length of TPSFLAT. The error decreases as sequence length increases. The largest maximum error is less than 0.22\%.}
    \vspace{-5pt}
    \label{fig:Figure8}
\end{figure}

\vspace{1em}
\textbf{Accuracy Stability |} We adopt the results of attention implemented in PyTorch with float data type as the baseline and simulate the behavior of FastTPS attention using bfloat16. The maximum discrepancy between FastTPS and the baseline is illustrated in Figure~\ref{fig:Figure8}. Our analysis reveals that the accuracy of FastTPS improves with increasing sequence length as the `Softmax` outputs become more uniform, causing the maximum value of V to approach 0.5. In contrast, FlashAttention exhibits a significant accuracy degradation as sequence length grows in Ref.~\cite{ref24}. Moreover, FastTPS achieves an order-of-magnitude higher accuracy ($\sim$$1e^{-3}$) compared to FlashAttention ($\sim$$1e^{-2}$)~\cite{ref25}. Notably, the primary source of precision loss in FastTPS stems from the truncation of output data rather than the FastTPS algorithm itself.

\section{DISCUSS AND FUTURE WORKS}
\subsection{Compatibility with Other Hardware}
In this work, we evaluate FastTPS on an NPU, though the framework is readily adaptable to GPUs, TPUs~\cite{ref60}, and ASICs~\cite{ref61, ref62, ref63, ref64}. Our attention mechanism builds upon FLAT~\cite{ref30} which has been validated on NVIDIA Tesla T4, and the KV Cache management and tiling strategies can be directly integrated without additional computational overhead. Meanwhile, our MLP dataflow optimization primarily involves storage reorganization. Given the shared memory hierarchy and the hardware architecture with GPUs, our approach is expected to yield performance gains on GPUs. Similarly, other AI accelerators—such as the DianNao family~\cite{ref61, ref62, ref63, ref64} and TPUs~\cite{ref60}—can achieve substantial performance improvements when integrated with FastTPS, despite their complex memory hierarchies which can be abstracted into a three-level cache structure (L1/L2/L3). Future work will explore broader applicability across different accelerators.

\subsection{Compatibility with Acceleration Algorithm}
As discussed in the introduction, numerous studies have explored Transformer model compression techniques, including quantization and distillation. Notably, FastTPS introduces low precision loss and remains fully compatible with these optimization methods, enabling seamless integration without performance trade-offs. This exceptional compatibility facilitates straightforward adoption across popular deep learning frameworks such as PyTorch and TensorFlow, allowing developers to easily extend these platforms with FastTPS capabilities. We plan to integrate these algorithms with FastTPS in the future.

\section{CONCLUSION}
This work tackles the pressing issue of low parallelism and low data usage efficiency of Transformer token phase on the general AI accelerators, especially for long sequence length. By introducing FastTPS with layout-aware KV Cache management, data tiling for reusing and fine-grain pipeline scheduling in both attention block and MLP block, we improved attention OI by 3x and brought MLP closer to the theoretical limit of bandwidth usage (1.3x than original data). The method shows strong compatibility with recent compression techniques such as GQA. Besides, FastTPS can be implemented on a variety of models and achieve more than 6x acceleration with little loss of accuracy, unlocking a practical route for deploying resource-efficient Transformer inference.


\bibliography{example_paper}

@article{ref1,
  title={A survey of GPT-3 family large language models including ChatGPT and GPT-4},
  author={Kalyan, Katikapalli Subramanyam},
  journal={Natural Language Processing Journal},
  volume={6},
  pages={100048},
  year={2024},
  publisher={Elsevier}
}

@inproceedings{ref2,
  title={Bert: Pre-training of deep bidirectional transformers for language understanding},
  author={Devlin, Jacob and Chang, Ming-Wei and Lee, Kenton and Toutanova, Kristina},
  booktitle={Proceedings of the 2019 conference of the North American chapter of the association for computational linguistics: human language technologies, volume 1 (long and short papers)},
  pages={4171--4186},
  year={2019}
}

@article{ref3,
  title={Attention is all you need},
  author={Ashish, Vaswani},
  journal={Advances in neural information processing systems},
  volume={30},
  pages={I},
  year={2017}
}

@article{ref4,
  title={Chatgpt: Optimizing language models for dialogue},
  author={Schulman, John and Zoph, Barret and Kim, Christina and Hilton, Jacob and Menick, Jacob and Weng, Jiayi and Uribe, Juan Felipe Ceron and Fedus, Liam and Metz, Luke and Pokorny, Michael and others},
  journal={OpenAI blog},
  volume={2},
  number={4},
  year={2022}
}

@article{ref5,
  title={Lamda: Language models for dialog applications},
  author={Thoppilan, Romal and De Freitas, Daniel and Hall, Jamie and Shazeer, Noam and Kulshreshtha, Apoorv and Cheng, Heng-Tze and Jin, Alicia and Bos, Taylor and Baker, Leslie and Du, Yu and others},
  journal={arXiv preprint arXiv:2201.08239},
  year={2022}
}

@article{ref6,
  title={Challenges and applications of large language models},
  author={Kaddour, Jean and Harris, Joshua and Mozes, Maximilian and Bradley, Herbie and Raileanu, Roberta and McHardy, Robert},
  journal={arXiv preprint arXiv:2307.10169},
  year={2023}
}

@article{ref7,
  title={End-edge-cloud collaborative computing for deep learning: A comprehensive survey},
  author={Wang, Yingchao and Yang, Chen and Lan, Shulin and Zhu, Liehuang and Zhang, Yan},
  journal={IEEE Communications Surveys \& Tutorials},
  volume={26},
  number={4},
  pages={2647--2683},
  year={2024},
  publisher={IEEE}
}

@inproceedings{ref8,
  title={Ray: A distributed framework for emerging $\{$AI$\}$ applications},
  author={Moritz, Philipp and Nishihara, Robert and Wang, Stephanie and Tumanov, Alexey and Liaw, Richard and Liang, Eric and Elibol, Melih and Yang, Zongheng and Paul, William and Jordan, Michael I and others},
  booktitle={13th USENIX symposium on operating systems design and implementation (OSDI 18)},
  pages={561--577},
  year={2018}
}

@article{ref9,
  title={Megatron-lm: Training multi-billion parameter language models using model parallelism},
  author={Shoeybi, Mohammad and Patwary, Mostofa and Puri, Raul and LeGresley, Patrick and Casper, Jared and Catanzaro, Bryan},
  journal={arXiv preprint arXiv:1909.08053},
  year={2019}
}

@inproceedings{ref10,
  title={Deepspeed: System optimizations enable training deep learning models with over 100 billion parameters},
  author={Rasley, Jeff and Rajbhandari, Samyam and Ruwase, Olatunji and He, Yuxiong},
  booktitle={Proceedings of the 26th ACM SIGKDD international conference on knowledge discovery \& data mining},
  pages={3505--3506},
  year={2020}
}

@inproceedings{ref11,
  title={$\{$ServerlessLLM$\}$:$\{$Low-Latency$\}$ serverless inference for large language models},
  author={Fu, Yao and Xue, Leyang and Huang, Yeqi and Brabete, Andrei-Octavian and Ustiugov, Dmitrii and Patel, Yuvraj and Mai, Luo},
  booktitle={18th USENIX Symposium on Operating Systems Design and Implementation (OSDI 24)},
  pages={135--153},
  year={2024}
}

@inproceedings{ref12,
  title={Artificial inteIligence on personal computers},
  author={Brodwin, David R and Erickson, Wayne and Kaplan, Jerrold and Rappaport, Wanda},
  booktitle={Managing Requirements Knowledge, International Workshop on},
  pages={461--461},
  year={1985},
  organization={IEEE Computer Society}
}

@inproceedings{ref13,
  title={Ai benchmark: Running deep neural networks on android smartphones},
  author={Ignatov, Andrey and Timofte, Radu and Chou, William and Wang, Ke and Wu, Max and Hartley, Tim and Van Gool, Luc},
  booktitle={Proceedings of the European Conference on Computer Vision (ECCV) Workshops},
  pages={0--0},
  year={2018}
}

@article{ref14,
  title={Understanding Large Language Models in Your Pockets: Performance Study on COTS Mobile Devices},
  author={Xiao, Jie and Huang, Qianyi and Chen, Xu and Tian, Chen},
  journal={arXiv preprint arXiv:2410.03613},
  year={2024}
}

@inproceedings{ref15,
  title={Splitwise: Efficient generative llm inference using phase splitting},
  author={Patel, Pratyush and Choukse, Esha and Zhang, Chaojie and Shah, Aashaka and Goiri, {\'I}{\~n}igo and Maleki, Saeed and Bianchini, Ricardo},
  booktitle={2024 ACM/IEEE 51st Annual International Symposium on Computer Architecture (ISCA)},
  pages={118--132},
  year={2024},
  organization={IEEE}
}

@article{ref16,
  title={Formal algorithms for transformers},
  author={Phuong, Mary and Hutter, Marcus},
  journal={arXiv preprint arXiv:2207.09238},
  year={2022}
}

@article{ref17,
  title={Kvquant: Towards 10 million context length llm inference with kv cache quantization},
  author={Hooper, Coleman and Kim, Sehoon and Mohammadzadeh, Hiva and Mahoney, Michael W and Shao, Yakun S and Keutzer, Kurt and Gholami, Amir},
  journal={Advances in Neural Information Processing Systems},
  volume={37},
  pages={1270--1303},
  year={2024}
}

@article{ref18,
  title={Q-vit: Accurate and fully quantized low-bit vision transformer},
  author={Li, Yanjing and Xu, Sheng and Zhang, Baochang and Cao, Xianbin and Gao, Peng and Guo, Guodong},
  journal={Advances in neural information processing systems},
  volume={35},
  pages={34451--34463},
  year={2022}
}

@article{ref19,
  title={Zeroquant: Efficient and affordable post-training quantization for large-scale transformers},
  author={Yao, Zhewei and Yazdani Aminabadi, Reza and Zhang, Minjia and Wu, Xiaoxia and Li, Conglong and He, Yuxiong},
  journal={Advances in neural information processing systems},
  volume={35},
  pages={27168--27183},
  year={2022}
}

@article{ref20,
  title={Vision transformer pruning},
  author={Zhu, Mingjian and Tang, Yehui and Han, Kai},
  journal={arXiv preprint arXiv:2104.08500},
  year={2021}
}

@article{ref21,
  title={A fast post-training pruning framework for transformers},
  author={Kwon, Woosuk and Kim, Sehoon and Mahoney, Michael W and Hassoun, Joseph and Keutzer, Kurt and Gholami, Amir},
  journal={Advances in Neural Information Processing Systems},
  volume={35},
  pages={24101--24116},
  year={2022}
}

@article{ref22,
  title={Flashattention: Fast and memory-efficient exact attention with io-awareness},
  author={Dao, Tri and Fu, Dan and Ermon, Stefano and Rudra, Atri and R{\'e}, Christopher},
  journal={Advances in neural information processing systems},
  volume={35},
  pages={16344--16359},
  year={2022}
}

@article{ref23,
  title={Flashattention-2: Faster attention with better parallelism and work partitioning},
  author={Dao, Tri},
  journal={arXiv preprint arXiv:2307.08691},
  year={2023}
}

@article{ref24,
  title={Flashattention-3: Fast and accurate attention with asynchrony and low-precision},
  author={Shah, Jay and Bikshandi, Ganesh and Zhang, Ying and Thakkar, Vijay and Ramani, Pradeep and Dao, Tri},
  journal={Advances in Neural Information Processing Systems},
  volume={37},
  pages={68658--68685},
  year={2024}
}

@article{ref25,
  title={Is flash attention stable?},
  author={Golden, Alicia and Hsia, Samuel and Sun, Fei and Acun, Bilge and Hosmer, Basil and Lee, Yejin and DeVito, Zachary and Johnson, Jeff and Wei, Gu-Yeon and Brooks, David and others},
  journal={arXiv preprint arXiv:2405.02803},
  year={2024}
}

@article{ref26,
  title={Efficientformer: Vision transformers at mobilenet speed},
  author={Li, Yanyu and Yuan, Geng and Wen, Yang and Hu, Ju and Evangelidis, Georgios and Tulyakov, Sergey and Wang, Yanzhi and Ren, Jian},
  journal={Advances in Neural Information Processing Systems},
  volume={35},
  pages={12934--12949},
  year={2022}
}

@article{ref27,
  title={Long-short transformer: Efficient transformers for language and vision},
  author={Zhu, Chen and Ping, Wei and Xiao, Chaowei and Shoeybi, Mohammad and Goldstein, Tom and Anandkumar, Anima and Catanzaro, Bryan},
  journal={Advances in neural information processing systems},
  volume={34},
  pages={17723--17736},
  year={2021}
}

@article{ref28,
  title={Improved precision and recall metric for assessing generative models},
  author={Kynk{\"a}{\"a}nniemi, Tuomas and Karras, Tero and Laine, Samuli and Lehtinen, Jaakko and Aila, Timo},
  journal={Advances in neural information processing systems},
  volume={32},
  year={2019}
}

@article{ref29,
  title={Ffn fusion: Rethinking sequential computation in large language models},
  author={Bercovich, Akhiad and Dabbah, Mohammad and Puny, Omri and Galil, Ido and Geifman, Amnon and Geifman, Yonatan and Golan, Izhak and Karpas, Ehud and Levy, Itay and Moshe, Zach and others},
  journal={arXiv preprint arXiv:2503.18908},
  year={2025}
}

@inproceedings{ref30,
  title={Flat: An optimized dataflow for mitigating attention bottlenecks},
  author={Kao, Sheng-Chun and Subramanian, Suvinay and Agrawal, Gaurav and Yazdanbakhsh, Amir and Krishna, Tushar},
  booktitle={Proceedings of the 28th ACM International Conference on Architectural Support for Programming Languages and Operating Systems, Volume 2},
  pages={295--310},
  year={2023}
}

@article{ref31,
  title={Amd xdna™ npu in ryzen™ ai processors},
  author={Rico, Alejandro and Pareek, Satyaprakash and Cabezas, Javier and Clarke, David and Ozgul, Baris and Barat, Francisco and Fu, Yao and M{\"u}nz, Stephan and Stuart, Dylan and Schlangen, Patrick and others},
  journal={IEEE Micro},
  year={2024},
  publisher={IEEE}
}

@inproceedings{ref32,
  title={AMD Ryzen™ 7040 series: Technology overview},
  author={Subramon, Mahesh and Kramer, David and Paul, Indrani},
  booktitle={2023 IEEE Hot Chips 35 Symposium (HCS)},
  pages={1--27},
  year={2023},
  organization={IEEE Computer Society}
}

@article{ref33,
  title={Forecasting LLM Inference Performance via Hardware-Agnostic Analytical Modeling},
  author={Patwari, Rajeev and Sirasao, Ashish and Das, Devleena},
  journal={arXiv preprint arXiv:2508.00904},
  year={2025}
}

@article{ref34,
  title={Chatglm: A family of large language models from glm-130b to glm-4 all tools},
  author={GLM, Team and Zeng, Aohan and Xu, Bin and Wang, Bowen and Zhang, Chenhui and Yin, Da and Zhang, Dan and Rojas, Diego and Feng, Guanyu and Zhao, Hanlin and others},
  journal={arXiv preprint arXiv:2406.12793},
  year={2024}
}

@article{ref35,
  title={Llama 2: Open foundation and fine-tuned chat models},
  author={Touvron, Hugo and Martin, Louis and Stone, Kevin and Albert, Peter and Almahairi, Amjad and Babaei, Yasmine and Bashlykov, Nikolay and Batra, Soumya and Bhargava, Prajjwal and Bhosale, Shruti and others},
  journal={arXiv preprint arXiv:2307.09288},
  year={2023}
}

@article{ref36,
  title={The llama 3 herd of models},
  author={Grattafiori, Aaron and Dubey, Abhimanyu and Jauhri, Abhinav and Pandey, Abhinav and Kadian, Abhishek and Al-Dahle, Ahmad and Letman, Aiesha and Mathur, Akhil and Schelten, Alan and Vaughan, Alex and others},
  journal={arXiv preprint arXiv:2407.21783},
  year={2024}
}

@article{ref37,
  title={Phi-4 technical report},
  author={Abdin, Marah and Aneja, Jyoti and Behl, Harkirat and Bubeck, S{\'e}bastien and Eldan, Ronen and Gunasekar, Suriya and Harrison, Michael and Hewett, Russell J and Javaheripi, Mojan and Kauffmann, Piero and others},
  journal={arXiv preprint arXiv:2412.08905},
  year={2024}
}

@misc{ref38,
  author = {Hugging Face},
  title = {Hugging Face – The AI community building the future},
}

@article{ref39,
  title={Transformer feed-forward layers are key-value memories},
  author={Geva, Mor and Schuster, Roei and Berant, Jonathan and Levy, Omer},
  journal={arXiv preprint arXiv:2012.14913},
  year={2020}
}

@article{ref40,
  title={Keep the cost down: A review on methods to optimize LLM's KV-cache consumption},
  author={Shi, Luohe and Zhang, Hongyi and Yao, Yao and Li, Zuchao and Zhao, Hai},
  journal={arXiv preprint arXiv:2407.18003},
  year={2024}
}

@article{ref41,
  title={Sigmoid-weighted linear units for neural network function approximation in reinforcement learning},
  author={Elfwing, Stefan and Uchibe, Eiji and Doya, Kenji},
  journal={Neural networks},
  volume={107},
  pages={3--11},
  year={2018},
  publisher={Elsevier}
}

@article{ref42,
  title={Llmem: Estimating gpu memory usage for fine-tuning pre-trained llms},
  author={Kim, Taeho and Wang, Yanming and Chaturvedi, Vatshank and Gupta, Lokesh and Kim, Seyeon and Kwon, Yongin and Ha, Sangtae},
  journal={arXiv preprint arXiv:2404.10933},
  year={2024}
}

@inproceedings{ref43,
  title={Dynamollm: Designing llm inference clusters for performance and energy efficiency},
  author={Stojkovic, Jovan and Zhang, Chaojie and Goiri, {\'I}{\~n}igo and Torrellas, Josep and Choukse, Esha},
  booktitle={2025 IEEE International Symposium on High Performance Computer Architecture (HPCA)},
  pages={1348--1362},
  year={2025},
  organization={IEEE}
}

@article{ref44,
  title={Memory is all you need: An overview of compute-in-memory architectures for accelerating large language model inference},
  author={Wolters, Christopher and Yang, Xiaoxuan and Schlichtmann, Ulf and Suzumura, Toyotaro},
  journal={arXiv preprint arXiv:2406.08413},
  year={2024}
}

@inproceedings{ref48,
  title={Efficient memory management for large language model serving with pagedattention},
  author={Kwon, Woosuk and Li, Zhuohan and Zhuang, Siyuan and Sheng, Ying and Zheng, Lianmin and Yu, Cody Hao and Gonzalez, Joseph and Zhang, Hao and Stoica, Ion},
  booktitle={Proceedings of the 29th symposium on operating systems principles},
  pages={611--626},
  year={2023}
}

@article{ref49,
  title={A comprehensive review of model compression techniques in machine learning},
  author={Dantas, Pierre Vilar and Sabino da Silva Jr, Waldir and Cordeiro, Lucas Carvalho and Carvalho, Celso Barbosa},
  journal={Applied Intelligence},
  volume={54},
  number={22},
  pages={11804--11844},
  year={2024},
  publisher={Springer}
}

@article{ref50,
  title={Deepseek-r1: Incentivizing reasoning capability in llms via reinforcement learning},
  author={Guo, Daya and Yang, Dejian and Zhang, Haowei and Song, Junxiao and Zhang, Ruoyu and Xu, Runxin and Zhu, Qihao and Ma, Shirong and Wang, Peiyi and Bi, Xiao and others},
  journal={arXiv preprint arXiv:2501.12948},
  year={2025}
}

@article{ref51,
  title={R-KV: Redundancy-aware KV Cache Compression for Training-Free Reasoning Models Acceleration},
  author={Cai, Zefan and Xiao, Wen and Sun, Hanshi and Luo, Cheng and Zhang, Yikai and Wan, Ke and Li, Yucheng and Zhou, Yeyang and Chang, Li-Wen and Gu, Jiuxiang and others},
  journal={arXiv preprint arXiv:2505.24133},
  year={2025}
}

@article{ref52,
  title={Can LLMs Maintain Fundamental Abilities under KV Cache Compression?},
  author={Liu, Xiang and Tang, Zhenheng and Chen, Hong and Dong, Peijie and Li, Zeyu and Zhou, Xiuze and Li, Bo and Hu, Xuming and Chu, Xiaowen},
  journal={arXiv preprint arXiv:2502.01941},
  year={2025}
}

@inproceedings{ref53,
  title={Insights into deepseek-v3: Scaling challenges and reflections on hardware for ai architectures},
  author={Zhao, Chenggang and Deng, Chengqi and Ruan, Chong and Dai, Damai and Gao, Huazuo and Li, Jiashi and Zhang, Liyue and Huang, Panpan and Zhou, Shangyan and Ma, Shirong and others},
  booktitle={Proceedings of the 52nd Annual International Symposium on Computer Architecture},
  pages={1731--1745},
  year={2025}
}

@article{ref56,
  title={Do you even need attention? a stack of feed-forward layers does surprisingly well on imagenet},
  author={Melas-Kyriazi, Luke},
  journal={arXiv preprint arXiv:2105.02723},
  year={2021}
}

@article{ref57,
  title={Llama: Open and efficient foundation language models},
  author={Touvron, Hugo and Lavril, Thibaut and Izacard, Gautier and Martinet, Xavier and Lachaux, Marie-Anne and Lacroix, Timoth{\'e}e and Rozi{\`e}re, Baptiste and Goyal, Naman and Hambro, Eric and Azhar, Faisal and others},
  journal={arXiv preprint arXiv:2302.13971},
  year={2023}
}

@article{ref58,
  title={Gqa: Training generalized multi-query transformer models from multi-head checkpoints},
  author={Ainslie, Joshua and Lee-Thorp, James and De Jong, Michiel and Zemlyanskiy, Yury and Lebr{\'o}n, Federico and Sanghai, Sumit},
  journal={arXiv preprint arXiv:2305.13245},
  year={2023}
}

@inproceedings{ref60,
  title={In-datacenter performance analysis of a tensor processing unit},
  author={Jouppi, Norman P and Young, Cliff and Patil, Nishant and Patterson, David and Agrawal, Gaurav and Bajwa, Raminder and Bates, Sarah and Bhatia, Suresh and Boden, Nan and Borchers, Al and others},
  booktitle={Proceedings of the 44th annual international symposium on computer architecture},
  pages={1--12},
  year={2017}
}

@inproceedings{ref61,
  title={ShiDianNao: Shifting vision processing closer to the sensor},
  author={Du, Zidong and Fasthuber, Robert and Chen, Tianshi and Ienne, Paolo and Li, Ling and Luo, Tao and Feng, Xiaobing and Chen, Yunji and Temam, Olivier},
  booktitle={Proceedings of the 42nd annual international symposium on computer architecture},
  pages={92--104},
  year={2015}
}

@article{ref62,
  title={Diannao: A small-footprint high-throughput accelerator for ubiquitous machine-learning},
  author={Chen, Tianshi and Du, Zidong and Sun, Ninghui and Wang, Jia and Wu, Chengyong and Chen, Yunji and Temam, Olivier},
  journal={ACM SIGARCH Computer Architecture News},
  volume={42},
  number={1},
  pages={269--284},
  year={2014},
  publisher={ACM New York, NY, USA}
}

@inproceedings{ref63,
  title={Dadiannao: A machine-learning supercomputer},
  author={Chen, Yunji and Luo, Tao and Liu, Shaoli and Zhang, Shijin and He, Liqiang and Wang, Jia and Li, Ling and Chen, Tianshi and Xu, Zhiwei and Sun, Ninghui and others},
  booktitle={2014 47th Annual IEEE/ACM International Symposium on Microarchitecture},
  pages={609--622},
  year={2014},
  organization={IEEE}
}

@article{ref64,
  title={Pudiannao: A polyvalent machine learning accelerator},
  author={Liu, Daofu and Chen, Tianshi and Liu, Shaoli and Zhou, Jinhong and Zhou, Shengyuan and Teman, Olivier and Feng, Xiaobing and Zhou, Xuehai and Chen, Yunji},
  journal={ACM SIGARCH Computer Architecture News},
  volume={43},
  number={1},
  pages={369--381},
  year={2015},
  publisher={ACM New York, NY, USA}
}

@misc{ref69,
  title        = {MLCommons Training Benchmarks},
  author       = {{MLCommons}},
  howpublished = {\url{https://mlcommons.org/benchmarks/training/}},
  note         = {Accessed February, 2026},
  year         = {2026}
}

@inproceedings{ref70,
  title        = {Lunar lake architecture session},
  author       = {Gihon, Arik},
  booktitle    = {2024 IEEE Hot Chips 36 Symposium (HCS)},
  organization = {IEEE Computer Society},
  year         = {2024}
}

@misc{ref71,
  title        = {AI Benchmarks Archive},
  author       = {{Jetson AI Lab}},
  howpublished = {\url{https://www.jetson-ai-lab.com/archive/benchmarks.html}},
  note         = {Accessed February, 2026},
  year         = {2026}
}

@misc{ref72,
  title        = {LFM-2 Benchmark Results},
  author       = {{FastFlow Labs}},
  howpublished = {\url{https://fastflowlm.com/docs/benchmarks/lfm2_results/}},
  note         = {Accessed February, 2026},
  year         = {2026}
}
\bibliographystyle{mlsys2025}

\appendix
\section{TPSFLAT Detailed Computation Process after Tiling}
\label{appendix:A1}
TPSFLAT adopts a two-level tiling strategy: an outer loop and an inner loop. The outer loop corresponds to data tiling from the L3 to the L2, while the inner loop handles tiling from the L2 to the L1. In this paper, we define the computation performed in each iteration at the L1 level as an ‘L1 tile,’ at the L2 level as an ‘L2 tile,’ and at the L3 level as an ‘L3 tile’ which corresponds to the inference task for a single token. As described in Algorithm 3, the outer loop initially partitions the computation along the \(H\) dimension, dividing all \(H\) heads into smaller groups of \(H_b\). The inner loop further splits each \(H_b\) group into individual head-level computations. Ultimately, each core completes the full attention computation for a single head’s data within the L1 cache.

\textbf{Steps 2--4.} In the initial stage of data transfer from $L_3$ to $L_2$, the tensors $Q$, $K$, $V_{cache}$, $sin$, and $cos$ are partitioned along the $H$ dimension into smaller blocks of size $H_b$. This results in corresponding sub-tensors: $Q_b$, $K_b$, $V_{cache,b}$, $sin_b$, and $cos_b$.

\begin{equation}
\frac{H}{H_b} \times (Q_b, K_b) = \text{Tiling\_L3\_Tile}(Q, K)
\label{eq:15}
\end{equation}

\begin{equation}
\text{where } (Q_b, K_b) \in \mathbb{R}^{\frac{H}{H_b} \times (B \times N \times H_b \times d)}
\end{equation}

\begin{equation}
\frac{H}{H_b} \times V_{cache_b} = \text{Tiling\_L3\_Tile}(V_{cache_b})
\label{eq:16}
\end{equation}

\begin{equation}
\text{where } V_{cache_b} \in \mathbb{R}^{\frac{H}{H_b} \times (B \times H_b \times N_t \times d)}
\end{equation}

\begin{equation}
\frac{H}{H_b} \times (sin_b, cos_b) = \text{Tiling\_L3\_Tile}(sin, cos)
\label{eq:17}
\end{equation}

\textbf{Steps 5--7.} Subsequently, during the transfer from the L2 to the L1, the data is further partitioned along the H dimension, with each block corresponding to a single head (i.e., H=1). This results in the sub-tensors $Q_i$, $K_i$, $V_{cache_i}$, $sin_i$, and $cos_b$. Notably, $Q_i$ and $K_i$ are stored in contiguous buffers upon being loaded into the L1. As a result, they can be treated as a single unified tensor, denoted as $QK_i$:

\begin{equation}
H_b \times (Q_i, K_i) = \text{Tiling\_L2\_Tile}(Q_b, K_b)
\label{eq:18}
\end{equation}
\begin{equation}
(Q_i, K_i) \in \mathbb{R}^{H_b \times (B \times N \times 1 \times d)}
\end{equation}

\begin{equation}
H_b \times V_{cache_i} = \text{Tiling\_L2\_Tile}(V_{cache_b})
\label{eq:19}
\end{equation}
\begin{equation}
V_{cache_i} \in \mathbb{R}^{H_b \times (B \times 1 \times N_t \times d)}
\end{equation}

\begin{equation}
H_b \times (sin_i, cos_i) = \text{Tiling\_L2\_Tile}(sin_b, cos_b)
\label{eq:20}
\end{equation}
\begin{equation}
(sin_i, cos_i) \in \mathbb{R}^{H_b \times (1 \times \frac{d}{2})}
\end{equation}

\begin{equation}
Q_i, K_i \rightarrow QK_i
\label{eq:21}
\end{equation}
\begin{equation}
QK_i \in \mathbb{R}^{2 \times (B \times N \times (H=1) \times d)}
\end{equation}

\textbf{Step 8.} Next, the sequence of fused kernel computations begins on the L1 cache. The first operation is the \texttt{RoPE} operator which produces the output tensor $QK_{i\_rope}$. Since the input to \texttt{RoPE} is the concatenated vector $QK_i$, composed of $Q_i$ and $K_i$, the output can be directly interpreted as two physically contiguous tensors: $Q_{i\_rope}$ and $K_{i\_rope}$. Furthermore, because $H=1$, the $H$ and $N$ dimensions can be interchanged without altering the data layout.

\begin{equation}
QK_{i\_rope} = \text{Rope}(QK_i, sin_i, cos_i) \rightarrow Q_{i\_rope}, K_{i\_rope}
\label{eq:22}
\end{equation}
\begin{equation}
(Q_{i\_rope}, K_{i\_rope}) \in \mathbb{R}^{B \times N \times (H=1) \times d}
\end{equation}

\begin{equation}
Q_{i\_rope}, K_{i\_rope} \rightarrow \mathbb{R}^{B \times (H=1) \times N \times d}
\label{eq:23}
\end{equation}

\textbf{Steps 9--10.} Subsequently, the tensor $K_{i\_rope}$ is written back from the L1 to the L3 via the L2, updating the KV Cache and producing the new $K\_cache$. Then, the corresponding $K_{cache\_i}$ is fetched from the L3 back into the L1. This tensor is then used together with $Q_{i\_rope}$ to perform the \texttt{QKT} computation, resulting in the intermediate attention score tensor $S_i$:

\begin{equation}
\text{L3} \rightarrow \text{L2} \rightarrow \text{L1}: K_{cache_i} \in \mathbb{R}^{B \times (H=1) \times N_t \times d}
\label{eq:24}
\end{equation}

\begin{equation}
S_i = \text{QKT}(Q_{i\_rope}, K_{cache_i})
\label{eq:25}
\end{equation}
\begin{equation}
S_i \in \mathbb{R}^{B \times (H=1) \times N \times N_t}
\end{equation}

\begin{equation}
P_i = \text{Softmax}(S_i)
\label{eq:26}
\end{equation}
\begin{equation}
P_i \in \mathbb{R}^{B \times (H=1) \times N \times N_t}
\end{equation}

\textbf{Step 12.} Next, the tensor $V_{cache\_i}$ is retrieved from the KV Cache in the L3 and loaded into the L1. It is then used together with $P_i$ to perform the \texttt{PV} computation, producing the output tensor $O_i$ which is subsequently written back to the L3 via the L2:

\begin{equation}
\text{L2} \rightarrow \text{L1}: V_{cache_i} \in \mathbb{R}^{B \times (H=1) \times N_t \times d}
\label{eq:27}
\end{equation}

\begin{equation}
O_i = \text{PV}(P_i, V_{cache_i})
\label{eq:28}
\end{equation}
\begin{equation}
O_i \in \mathbb{R}^{B \times (H=1) \times N \times d}
\end{equation}

\textbf{Steps 13--14.} Finally, the iterative computation is completed. The intermediate outputs $O_i$ are first aggregated within the L2 to form the partial result $O_b$. Subsequently, these results are further accumulated and propagated to the L3, yielding the final output $O$ of the attention block.


\section{Cache Usage in FASTFLAT}
\label{appendix:A2}

Given that the cache sizes decrease progressively from the L3 to the L2 and then to the L1, we adopt a two-level tiling strategy: from the L3 to the L2, and from the L2 to the L1. This hierarchical tiling design provides strong adaptability, allowing the approach to accommodate variations in both hardware resources and model sizes by adjusting the tiling granularity at each level. Specifically, the tiling strategy can be parameterized in terms of the buffer requirements at each cache level. In our scheme, the granularity of partitioning along the $H$ dimension determines the buffer occupancy in the L1 and L2. As shown in Table~\ref{table-cache-usage}, the table summarizes the data volume required at each cache level: L3 cache must store the full input and output token tensors $O(BND)$, the KV Cache $O(BN_t D)$, and the sinusoidal components required for \texttt{RoPE} $O(BN_t D)$. The L2 cache holds sub-blocks of the input and output token tensors $O(H_b Nd)$, sub-blocks of the KV Cache $O(H_b N_t d)$, and the corresponding \texttt{RoPE} sine and cosine sub-blocks $O(H_b d)$. The L1 cache requires further partitioning of the input/output tokens, KV Cache, and \texttt{RoPE}-related data, with corresponding data volumes of $O(Nd)$, $O(N_t d)$, and $O(d)$, respectively. Additionally, temporary intermediate variables are needed, with a data volume of $O(NN_t)$.

\begin{table}[t]
\caption{Cache usage across different levels.}
\label{table-cache-usage}
\begin{center}
\resizebox{\linewidth}{!}{%
\begin{tabular}{lccc}
\toprule
Cache Type & L1 & L2 & L3 \\
\midrule
Buffer Req. & 
$O(Nd) + N_t d + NN_t + d$ & 
$O(H_b Nd) + H_b N_t d + H_b d$ & 
$O(BND) + BN_t D + Hd$ \\
\bottomrule
\end{tabular}
}
\end{center}
\vskip -0.1in
\end{table}

\section{MLP Implementation Details}
\label{appendix:A3}
\textbf{Steps 1--4.} These steps outline the data transfer strategies from the L3 cache to the L2 cache of the MLP tiling. Each iteration can load $\bm{h}_{t}^{nkd} \in \mathbb{R}^{(x*tile_n),x(z*tile_d)}$ and $\bm{W}_{gu}^{nkd} \in \mathbb{R}^{(z*tile_d),(y*z*tile_k)}$ from the L3 cache to the L2 cache through the outer loop approach.

Steps 5--9: These steps outline the data transfer strategies from the L2 cache to the L1 cache of the MLP tiling. Each iteration can load $\bm{h}_{t}^{i} \in \mathbb{R}^{tile_n \times tile_d}$
and $\bm{W}_{gu}^{k_i} \in \mathbb{R}^{tile_d \times tile_k}$ from the L2 cache to the L1 cache through the inner-loop approach. This step pertains to the gate-up projection fusion part of the MLP tiling strategy. It performs matrix multiplication on the tile blocks loaded into L1 cache by Step~7, then accumulates the results in the inner loop to obtain $\bm{h}_{t}^{gu_i} \in \mathbb{R}^{tile_n \times tile_k}.$

\begin{equation}
\bm{h}_{t}^{gu_i} = \sum_{count = 1}^{D / tile_D} \bm{w}_{gu}^{i} \, \bm{h}_{t}^{i}
\label{eq:29}
\end{equation}

Steps~10--11: These steps reduce all $\bm{h}_{t}^{gu_i}$ to get $\bm{h}_{t}^{gu_{nk}}$, followed by the fused ``siLu'' multiplication part of the MLP tiling strategy. The tensor $\bm{h}_{t}^{gu_{nk}} \in \mathbb{R}^{x \times tile_n \times y \times tile_k}$ is processed by the fused operator ``siLu\_Mul'' after inner-loop accumulation to obtain $\bm{h}_{t}^{sm_{nk}} \in \mathbb{R}^{x \times tile_n \times y \times tile_k}$, as shown in Equation~(30). Concretely, we first compute the ``siLu'' of the first half of the gate tensor $\bm{h}_{t}^{g_{nk}} \in \mathbb{R}^{x \times tile_n \times [0,\, y \cdot tile_k/2]}$, then perform elementwise multiplication with the other half from the up part $\bm{h}_{t}^{u_{nk}} \in \mathbb{R}^{x \times tile_n \times [y \cdot tile_k/2,\, y \cdot tile_k]}$ to obtain the fused ``siLu\_Mul'' result, as described in Equation~(31).

\begin{equation}
\bm{h}_{t}^{\mathrm{sm}_{nk}} = \mathrm{siLu}\_{\mathrm{Mul}}\!\left( \bm{h}_{t}^{\mathrm{gu}_{nk}} \right)
\label{eq:30}
\end{equation}

\begin{equation}
\bm{h}_{t}^{\mathrm{sm}_{nk}} 
= \mathrm{siLu}\!\left( \bm{h}_{t}^{g_{nk}} \right)
  \ast 
  \bm{h}_{t}^{u_{nk}}
\label{eq:31}
\end{equation}

Step~12: The step outlines the data transfer strategy from the L2 cache to the L3 cache of the MLP tiling. When the quantity of computational units in the L2 cache ready to be written back reaches the threshold for a single write-back operation to the L3 cache, the $\bm{h}_{t}^{sm_{nk}}$ units in the L2 will be written back to the L3 cache at one time.

Steps~13--15: These steps pertain to the down-projection part of the MLP tiling strategy. 
This section will not be elaborated in detail since this feature is outside the scope of our optimization.

\begin{equation}
\bm{O} = \bm{W}_{down}\, \bm{h}_{t}^{sm}
\label{eq:32}
\end{equation}

\section{Other benchmarks}
\label{appendix:A4}

\begin{table}[t]
\centering
\caption{Token per second for 3 models on AMD and Intel NPU respectively for MLPerf benchmark.}
\label{tab:mlperf_tps}
\small

\begin{tabular}{lccc}
\toprule
\textbf{Category (Llama3.2)} & \textbf{AMD TPS} & \textbf{LNL TPS} & \textbf{AMD/LNL} \\
\midrule
Content Gen            & 27.12 & 24.69 & 1.10 \\
Creative Writing       & 25.06 & 25.93 & 0.97 \\
Summarization, Light   & 23.34 & 22.28 & 1.05 \\
Summarization, Moderate& 22.27 & 19.83 & 1.12 \\
Code Generation        & 20.94 & 20.15 & 1.04 \\
\bottomrule
\end{tabular}

\vspace{0.6em}

\begin{tabular}{lccc}
\toprule
\textbf{Category (Llama2)} & \textbf{AMD TPS} & \textbf{LNL TPS} & \textbf{AMD/LNL} \\
\midrule
Content Gen            & 16.89 & 22.11 & 0.76 \\
Creative Writing       & 15.79 & 19.28 & 0.82 \\
Summarization, Light   & 14.94 & 17.16 & 0.87 \\
Summarization, Moderate& 14.48 & 17.08 & 0.85 \\
Code Generation        & 13.78 & 14.84 & 0.93 \\
\bottomrule
\end{tabular}

\vspace{0.6em}

\begin{tabular}{lccc}
\toprule
\textbf{Category (Llama3)} & \textbf{AMD TPS} & \textbf{LNL TPS} & \textbf{AMD/LNL} \\
\midrule
Content Gen            & 16.16 & 17.76 & 0.91 \\
Creative Writing       & 15.92 & 18.99 & 0.84 \\
Summarization, Light   & 15.15 & 19.97 & 0.76 \\
Summarization, Moderate& 15.47 & 19.33 & 0.80 \\
Code Generation        & 15.27 & 18.92 & 0.81 \\
\bottomrule
\end{tabular}

\end{table}

\begin{table}[t]
\centering
\caption{Input seq = 2048, output = 128, compared with Nvidia Nano (102\,GBPs) and AMD Ryzen AI 400 series Gorgon Point~2.}
\label{tab:table6}
\small
\begin{tabular}{lccc}
\toprule
\textbf{Category} & \textbf{AMD TPS} & \textbf{Nano TPS} & \textbf{AMD/Nano} \\
\midrule
Llama 3.1 8B  & 10.79 & 14.00 & 0.77 \\
Llama 3.2 3B  & 21.60 & 27.70 & 0.78 \\
Qwen2.5 7B    & 10.49 & 14.20 & 0.74 \\
\bottomrule
\end{tabular}
\end{table}

\begin{table}[t]
\centering
\caption{Max sequence = 1k/2k/4k of LFM2 1.2B/2.6B with FastFlowLM; AMD platform: Ryzen AI 400 series Krackan Point.}
\label{tab:table7}
\small

\begin{tabular}{lccc}
\toprule
\textbf{Max Sequence (LFM2 1.2B)} & \textbf{AMD TPS} & \textbf{FastFlowLM} & \textbf{AMD/FFL} \\
\midrule
1k  & 63.02 & 61.3 & 1.03 \\
2k  & 62.32 & 60.5 & 1.03 \\
4k  & 62.17 & 57.5 & 1.08 \\
\bottomrule
\end{tabular}

\vspace{0.7em}

\begin{tabular}{lccc}
\toprule
\textbf{Max Sequence (LFM2 2.6B)} & \textbf{AMD TPS} & \textbf{FastFlowLM} & \textbf{AMD/FFL} \\
\midrule
1k  & 32.87 & 30.3 & 1.08 \\
2k  & 32.73 & 29.9 & 1.09 \\
4k  & 31.96 & 29.0 & 1.10 \\
\bottomrule
\end{tabular}

\end{table}

FastTPS has been integrated into the AMD software stack and benchmarked against Intel’s solution using the MLPerf workload suite~\cite{ref69}. Evaluations were conducted on the AMD Ryzen AI 400 series Gorgon Point 2 NPU platform, which provides a sustained memory bandwidth of 80\,GBPs, and on the Intel Lunar Lake NPU platform~\cite{ref70}, which offers a bandwidth of 102\,GBPs. The comparative performance results are summarized in Table~5.

Besides, the speeds are compared with the Nvidia Nano~\cite{ref71} and FastFlowLM~\cite{ref72} on specific prompt length and token length for some models, see Table~6 and Table~7.

\end{document}